\documentclass{article}
\usepackage[utf8]{inputenc}
\usepackage{graphicx}
\usepackage{algorithm2e}
\usepackage{mathtools}
\usepackage{subcaption}
\usepackage{multirow}

\title{Oil reservoir recovery factor assessment using Bayesian networks based on advanced approaches to analogues clustering}
\author{Petr Andriushchenko, Irina Deeva, Anna Bubnova, \\
Anton Voskresenskiy, Nikita Bukhanov, \\
Nikolay Nikitin and Anna Kalyuzhnaya \\
\\
ITMO University, Saint-Petersburg, Russia}
\date{March 2022}

\begin{document}

\maketitle

\begin{abstract}
The work focuses on the modelling and imputation of oil and gas reservoirs parameters, specifically, the problem of predicting the oil recovery factor (RF) using Bayesian networks (BNs). Recovery forecasting is critical for the oil and gas industry as it directly affects a company's profit. However, current approaches to forecasting the RF are complex and computationally expensive. In addition, they require vast amount of data and are difficult to constrain in the early stages of reservoir development. To address this problem, we propose a BN approach and describe ways to improve parameter predictions' accuracy. Various training hyperparameters for BNs were considered, and the best ones were used. The approaches of structure and parameter learning, data discretization and normalization, subsampling on analogues of the target reservoir, clustering of networks and data filtering were considered. Finally, a physical model of a synthetic oil reservoir was used to validate BNs' predictions of the RF. All approaches to modelling based on BNs provide full coverage of the confidence interval for the RF predicted by the physical model, but at the same time require less time and data for modelling, which demonstrates the possibility of using in the early stages of reservoirs development. The main result of the work can be considered the development of a methodology for studying the parameters of reservoirs based on Bayesian networks built on small amounts of data and with minimal involvement of expert knowledge. The methodology was tested on the example of the problem of the recovery factor imputation.
\end{abstract}

\section{Introduction}
Today, industry specialists are faced with various issues related to the analysis of large amounts of data from oil and gas reservoirs. Many of these issues can be efficiently solved using machine learning. For example, an important geologic task is to find analogous reservoirs \cite{popova2018analogy}. In terms of time and labour costs, the simplest method to search for reservoir analogues is to consider oil and gas reservoirs located near to the study area \cite{voskresenskiy2020variations}. If we consider searching for analogues as a clustering problem in a multidimensional space, it becomes possible to consider various clustering algorithms to solve the problem. In previous studies \cite{silva2018sensitivity, martin2013new}, hierarchical clustering has been proposed to search for analogues reservoirs. Another critical task is analysing and predicting geological parameters, such as Lithology, Porosity, Depositional Environment. This task can be solved using artificial neural networks, support vector machines, classification and regression models \cite{ani2016reservoir}. It was shown that machine learning methods could infer domain knowledge from well log data and identify the most important parameters in reservoir analogues data. Further studies  \cite{sircar2021application,thanh2020application, thanh2021integrated,mazumder2021failure} show other cases of application of machine learning methods in the oil and gas industry, including history matching, enhanced oil recovery and equipment failure risk analysis. These approaches usually solve a specific problem, while the oil and gas industry needs a flexible and multifunctional tool to solve a broad range of problems based on subsurface data.

If data on geological parameters are considered a multivariate distribution, then probabilistic graphical models, in particular Bayesian networks (BN), can become such a tool. These models allow rapidly obtaining causal relations between geological parameters and present these relations in an interpretable way. Contrary to deep learning methods and fluid flow physical modelling based on partial differential equations (PDEs), probabilistic graphical models present the results of structure and parameter learning based on a worldwide database of oil reservoirs in a form that can be intuitively managed and edited by domain experts. BNs are already used in the oil and gas industry, for example, to identify the significance of geological parameters \cite{masoudi2015feature, martinelli2013building}. BNs can also be used to reveal new knowledge about an object, fill in gaps, and find anomalous values \cite{deeva2021oil, andriushchenko2020analysis}. This paper discusses an approach based on the BN construction for the problem of modelling and predicting reservoir parameters (fig. \ref{pairplot}). 

\begin{figure}
\includegraphics[width=1\linewidth]{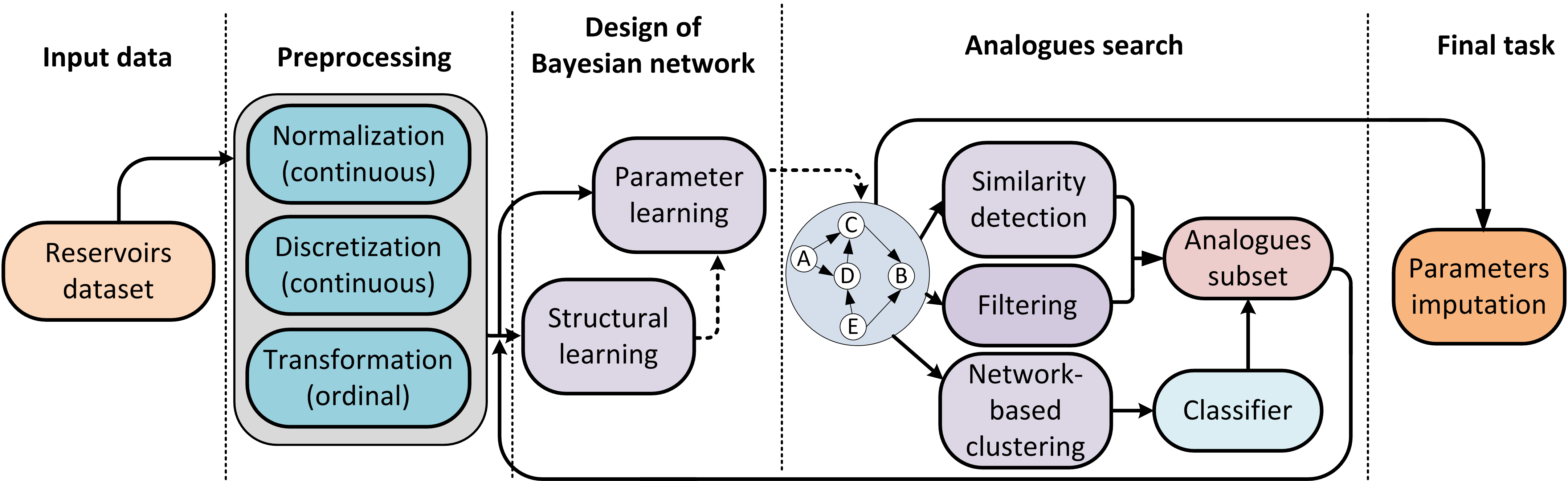}
\caption{General pipeline of the approach for modelling and predicting geological reservoir parameters based on BNs. \label{pipeline}}
\end{figure}

The general scheme of the approach is shown in the fig. \ref{pipeline}. It consists of several blocks: (1) data preprocessing, (2) structure and parameter learning, and (3) analogues search. The first block allows preparing the data for probabilistic modelling using different transformations. The second block generated the BNs that can represent the preprocessed data from the previous stage. Finally, the third block is aimed to analyze the similarity between reservoirs to improve the imputation quality for a specific subset of data. By the problem of imputation, we mean the prediction of values in a BN node based on values from parent nodes. As an example of practical application for the proposed approach, we consider a specific geological case -- the prediction of the recovery factor (RF) parameter.

In the first place, we included parameters that affect RF from a domain point of view, such as porosity, permeability, lithology, to the model to predict RF. Then we carried out how other parameters presented in the dataset affect RF prediction. Domain experts may not have taken into account these parameters since they do not usually consider them throughout their careers. Consequently, we have included parameters such as Tectonic regime, Structural setting, Period, Depth etc. As a result, we included parameters that could be analyzed and representative in the database to the model.

\subsection{Recovery factor}
The relevance of the proposed method for RF prediction is the quick assessment of a reservoir potential enhanced by causal inference analysis between reservoir parameters. RF is one of the most significant parameters for oil and gas companies during the appraisal and evaluation stage in order to support investment decisions based on limited information about reservoir \cite{noureldien2015using}. The RF is a productive portion of the hydrocarbon in place, typically ranging 20-40\% for oil and 50-80\% for gas \cite{makhotin2020machine, jahn2003hydrocarbon, muggeridge2013recovery}. Even a small increase in RF could add many extra barrels, strengthening the economic feasibility of reservoir development. Therefore, careful assessment of parameters that influence RF is a vital task.

RF estimation depends on many factors such as reservoir quality, properties of reservoir fluids, development strategy, producing time. Reservoir with higher reservoir quality in terms of porosity and permeability are characterized by higher RF because they contain a larger amount of hydrocarbons, and pore radii are higher, allowing fluids to flow more easily to producing wells. Low initial water saturation positively influences the RF as relative permeability of the hydrocarbon phase would be higher during a more significant time. Usually, RFs of gas reservoirs are much higher than oil RFs due to higher mobility (the productivity of a well is directly proportional to the mobility). Mobility is the ratio of effective permeability to the fluid phase viscosity. Reservoir compartmentalization impacts RF since reservoir volumes with movable fluid could be restricted by producing wells \cite{jolley2010reservoir}. A Depositional environment may influence recovery efficiency. For instance, less heterogeneous depositional environments, such as wave-dominated deltas, may demonstrate RFs more than 50\%. On the other hand, RFs of more complex environments, such as fluvial dominated deltas, lie within 20-40\% range \cite{tyler1991architectural, larue2005controversy}. Reservoir lithology can influence RF. For example, terrigenous reservoirs usually have higher RFs in comparison with carbonate ones \cite{sloan2003quantification, shepherd2009factors}. Factors that contribute significantly to RF are development strategy and reservoir drive mechanism. A relatively low RF will characterize an oil reservoir with a depletion drive mechanism. However, waterflooding (or other enhanced hydrocarbon recovery methods) could be drastically increased to obtain supplementary recovery. Enhanced recovery methods aim to increase the natural energy of the reservoir, usually by displacing the hydrocarbons towards the producing wells with some injected fluid \cite{dake1978fundamentsls}. Increasing reservoir producing time will increase the recovery, but mature reservoirs usually produce hydrocarbons with water, which should be separated from hydrocarbon and processed. So, production time is mainly determined by economic feasibility and water cut.

Our proposed method generates distributions of the RF in seconds (0.5 seconds for 500 samples), considering variations in input reservoir properties depending on the conceptual scenario. A distribution of RFs is needed to capture the uncertain nature of some reservoir parameters, such as Lithology, Porosity, Depositional Environment. The results can be used to evaluate oil or gas RFs especially in the early stages of development when data are likely to be scarce.

\subsection{Description of approaches to recovery factor estimation}

\begin{table}[!b]
\centering
\caption{Comparison of approaches to the RF estimation.
\label{table_comparison}}
\begin{tabular}{|p{0.16\textwidth}|p{0.11\textwidth}|p{0.11\textwidth}|p{0.11\textwidth}|p{0.11\textwidth}|p{0.11\textwidth}|p{0.10\textwidth}|}
\hline

\textbf{Approach} & \textbf{Applied in early stages of \newline reservoir life} & \textbf{Domain expert required} & \textbf{Subjec-tivity} & \textbf{Special software required} & \textbf{Model development speed*} & \textbf{Speed of recalculation when new data are received} \\ \hline

Reservoir analogues & + & + & + & - & days & days \\   
& \multicolumn{6}{c}{\parbox{10cm}{Depends on representative data availability, subject to inaccuracies in selecting analogues and biased towards the previous experience of an expert}} \vline\\ \hline

Empirical correlations and \newline benchmarking & + & + & + & - & days & days \\
& \multicolumn{6}{c}{\parbox{10cm}{Depends on representative data availability}} \vline\\ \hline

Decline curve \newline analysis and \newline material balance methods & +/- & + & +/- & + & days & hours/\linebreak days \\
& \multicolumn{6}{c}{\parbox{10cm}{Requires well test and production data}} \vline\\ \hline

Simulation of a 3D model & - & + & - & + & weeks & days \\
& \multicolumn{6}{c}{\parbox{10cm} {Incorporates both material balance and fluid flow equations}} \vline\\ \hline

Bayesian \newline network & + & - & - & - & 5 minutes** & 5 minutes** \\
& \multicolumn{6}{c}{\parbox{10cm}{Provides predictive uncertainty}} \vline\\ \hline

\end{tabular}
"+" and "-" denote the presence or absence of parameters defined in the table columns. "+/-" denotes the fact that it is hard to say for sure whether the method is applicable due to data availability issues or the method could be subjective in some cases. \newline * Model development speed was estimated by a group of domain experts.\newline
** Data cleaning and preparation did not take into account.
\end{table}

Generally, there are four main approaches to RF and reserves calculation. The first one is based on an expert's selection of reservoir analogues and an estimation of the range of possible fractions of recoverable hydrocarbons from a reservoir. To be more specific, an expert estimates the probability density function of RFs from analogues to constrain RF estimation on a target reservoir. Analogue selection may be biased by their representation in a database or the experience of an expert. The main advantage of this method, however, is that it may be used in the early stages of field development and could be based on different sets of reservoir parameters, which may be results of expert's interpretations (Lithology, Tectonic setting) or measured values of rocks and fluids (Porosity, Oil Density). The cons of the analogue method are 1) most often, domain experts consider analogues only nearby to a target reservoir and do not analyze world analogues, 2) there is no methodologies and best practices for selecting analogues, 3) retrospective analysis of picked analogues is not performed \cite{sloan2004global}.

The second method uses empirical correlations between reservoir parameters taking into account such parameters as porosity, permeability, fluid density, viscosity, reservoir heterogeneity and discontinuity, and other parameters of developed fields to predict the RF \cite{wickens2010rapid, jia2016novel, mahmoud2019estimation}. Usually, multivariate regression analysis techniques are used to develop such correlations \cite{gomes2018benchmarking}. In other words, a second method is a benchmarking approach of comparing the RF of a target reservoir with other reservoirs having comparable reservoir properties and development strategies. In that way, the method uses a specific equation that predicts what the RF should be, based on a set of given reservoir parameters. In some ways, it is similar to the first approach and, as a result, is subjective and requires the direct participation of a domain expert. Over time, RF efficiency increases due to new technologies and improvements to existing ones, but the first and second approaches do not take it into consideration. On the other hand, reservoir quality of developed fields has decreased over time \cite{oil2017recovery}. Therefore, reservoir complexity also increases over time, but this approach does not take it into consideration. Both previous approaches do not require special software, and they could be time-consuming as it highly depends on reservoir data parameters availability.

The third approach uses decline curve analysis and material balance methods \cite{busby2020deep, craft1992applied, ahmed2010reservoir}. These methods are widely used by experts in the industry but require specific data (sufficient fluid production data and data related to water encroachment from the reservoir), domain expertise and are time-consuming. The main idea of the decline curve analysis is an estimation of cumulative production by analyzing declining production rates and forecasting future reservoir performance. It could be evaluated from a chart of hydrocarbon production rate with time or from a plot of production rates versus cumulative hydrocarbon production \cite{rahuma2013prediction}. The hydrocarbon flow rate of the well or field decreases with increasing production time, provided there is no maintenance of reservoir pressure. The material balance technique considers a reservoir as a tank model and attempts to balance changes in reservoir volume as a result of production. However, the method could not be applicable for unconventional reservoirs \cite{al2016developed}. Decline curve analysis is usually performed by commercial software, but material balance methods can be carried out without any special programs. Material balance is considered a non-subjective method. It is very similar to simulation with some simplification. Decline curve analysis, on the other hand, in some cases may be subjective. Both approaches can be performed only when some production data are available.

The forth approach uses 3D reservoir simulation to estimate possible RFs \cite{craft1992applied, ahmed2010reservoir}. This method combines fluid flow and material balance equations. It is based on the physics of multiphase fluid flow and supports the fine-tuning of reservoir and production parameters, such as number and type of wells or well maximum flowrate limit. The main drawback is that building the model is time consuming and specialized simulation software is required. For example, one simulation could take up to a few hours, but hundreds of them may be required to estimate uncertainty in reservoir RF correctly. Reservoir simulation is preferred when a significant amount of reservoir information and production data are available and considered as a non-subjective method as it is based on production history data.

This paper proposes a different approach to modelling and predicting reservoir parameters (both interpretations and measurements). Probabilistic graphical models such as BNs support the discovery of implicit relationships in data. This approach enables identification of anomalies and prediction of reservoir parameters values for a fairly short period of time. We design our approach as an exploratory tool which widely uses help of geoscience experts. Pre-learning stage participation of expert include (1) detailed choice of learning algorithms, parameters to model and discretization methods, (2) direct inclusion and exclusion of particular edges of networks which are explicitly defined by expert and are taken into account during learning procedure, (3) filtering option which allows to create networks for particular basins, stratigraphy and ranges of parameters. Post-learning stage include: (1) usage of created networks as a monitoring tool for particular reservoir to be sure that all assumed relations are in place and validated by data obtained on a field, (2) rapid forecast of possible recovery factor and other parameters for a particular region with limited information, (3) comparison between different networks which allow to delineate robust relations across different geological settings. However, there are also downsides to this approach: since this model relies almost entirely on data, the BN requires good quality data for reservoirs parameters. The accuracy of the proposed approach depends on data availability. The model is also sensitive to data preprocessing and BNs learning approaches. That is why these approaches will be considered and compared in detail below.
A comparison of described approaches are presented in Table \ref{table_comparison}.

\section{Dataset description}

The training dataset consisted of 318 reservoirs. It was collected from open data sources and Gazpromneft's databases. The dataset contains reservoirs from all over the world. The dataset includes both categorical and continuous parameters of reservoirs. We considered the following parameters of the reservoirs, which were sufficiently represented for all reservoirs in our dataset:
\begin{enumerate}
    \item Tectonic Regime (categorical) represents the dominant type of processes which control structure and properties of strata and its dynamic evolution
    \item Structural Setting (categorical) describes combination of rock units with respect to their deformation histories
    \item Lithology (categorical) is a description of physical characteristics of the rock which include mineral composition, color, texture, size of the particles
    \item Porosity (continuous) is a measure of space within a rock that is typically filled with water, oil or gas. It is expressed as a fraction of the volume of voids over the total volume
    \item Permeability (continuous) is the property of rocks that refers to their ability to transfer fluids
    \item Depth (continuous) is a true vertical depth of the reservoir top that measured in meters below kelly bushing
    \item Gross (continuous) is the total vertical thickness of the reservoir
    \item Net Pay (continuous) is the vertical thickness of the reservoir that is considered to have adequate porosity to hold hydrocarbons
    \item Oil Density (continuous)  is a ratio between oil mass and its volume
    \item NTG (Sandiness = Net Pay/Gross) (continuous) is ration of Net Pay to Gross
    \item Oil RF (continuous) is the recoverable portion of oil initially in place, expressed as a percentage. Within the proposed workflow it is used for validation of constructed BNs
\end{enumerate}

\begin{figure}
\includegraphics[width=1\linewidth]{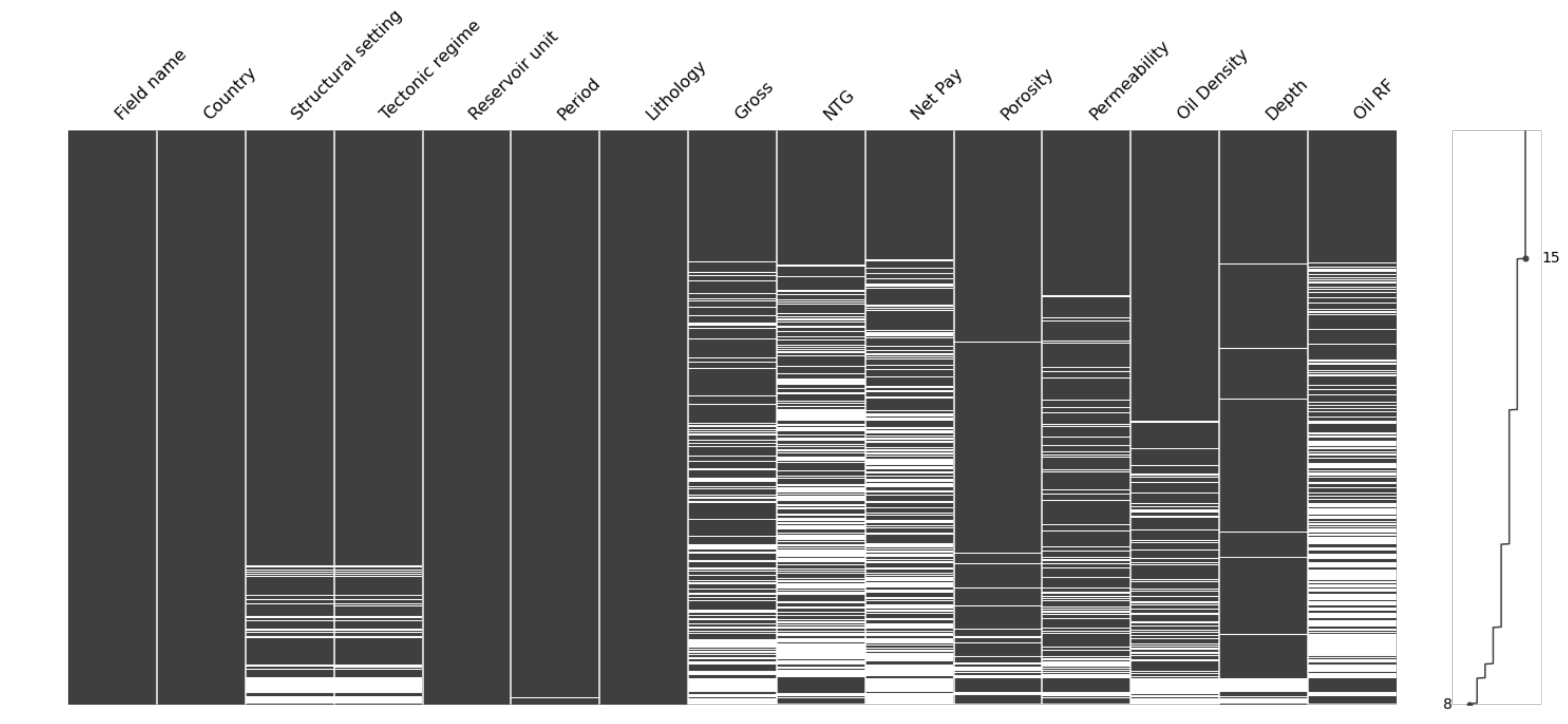}
\caption{Dataset completeness chart. The rightmost curve shows the number of columns in a row without missing values.  The gray and white colors on the graph represent the presence or absence of data in a particular column and row respectively.} \label{missing_values}
\end{figure}

Fig. \ref{missing_values} visualize data availability in the dataset. The reservoirs in this chart are ordered from most to least complete. The rightmost curve shows the number of columns in a row without missing values. On the figure you can see the upper part, in which there are no reservoirs with gaps in the considering parameters. The gray and white colors on the graph represent the presence or absence of data in a particular column and row, respectively.

Such a set of parameters is because to build BNs it is not necessary to select only those that affect the RF; on the contrary, the set of parameters should be expanded to obtain more complex and hierarchical dependencies. Apart from expert knowledge which would indicate already described and validated interconnections, BN would reveal potentially hidden relationship between parameters. This approach to finding relationships is interpreted by the initial view of the reservoir as a composite probabilistic object, that is, an object that is always described by various parameters that are related to each other and represent a multidimensional distribution.

When the database was collected, most of the categorical parameters were interpreted by the company's domain experts, such as petrophysicists and sedimentologists. Continuous variables are collected from a company's databases, and inherit some uncertainty as a result of aggregation. Categorical variables have some subjectivity since domain experts can interpret raw data slightly differently. Due to confidentiality, our dataset is not available, but a publicly available analogue of the dataset can be found in the repository \cite{BAMT}.

Categorical parameters include the following: ``Tectonic Regime'' has five different categories, ``Structural Setting'' has 12 different categories, ``Lithology'' has 13 different categories, and ``Period''. The age of reservoirs ranges from From Neogene to Cambrian. The values in categories are presented in Table \ref{table_list_values}.

\begin{table}
    \centering
    \caption{A list of unique values in categorical variables}
    \label{table_list_values}
    \begin{tabular}{|l|l|l|} 
        \hline
        \textbf{Tectonic Regime}& \textbf{Structural Setting} & \textbf{Lithology} \\ 
        \hline

        \parbox{0.2\textwidth}{
            \begin{description}
            \item Compression
            \item Extension
            \item Strike-slip
            \item Gravity
            \item Inversion
        \end{description}} & 
\parbox{0.3\textwidth}{
\begin{description}
    \item Intracratonic
    \item Rift
    \item Salt
    \item Inversion 
    \item Wrench
    \item Foreland
    \item Passive margin
    \item Sub-salt
    \item Delta
    \item Thrust
    \item Backarc
    \item Sub-thrust 
\end{description}}
&
\parbox{0.4\textwidth}{\begin{description}
    \item Limestone
    \item Chalky limestone
    \item Chalk
    \item Dolomite
    \item Dolomitic limestone
    \item Conglomerate
    \item Sandstone
    \item Low-resistivity Sandstone
    \item Thinly bedded Sandstone
    \item Shaly Sandstone
    \item Siltstone
    \item Basement
    \item Volcanics
\end{description}} \\

\hline
\end{tabular}
\end{table}

Some categorical parameters can be considered not as nominal but as ordinal. This issue will be discussed in more detail in the Section \ref{subsection_nominal_to_ordinal}.

\begin{figure}
\includegraphics[width=1\linewidth]{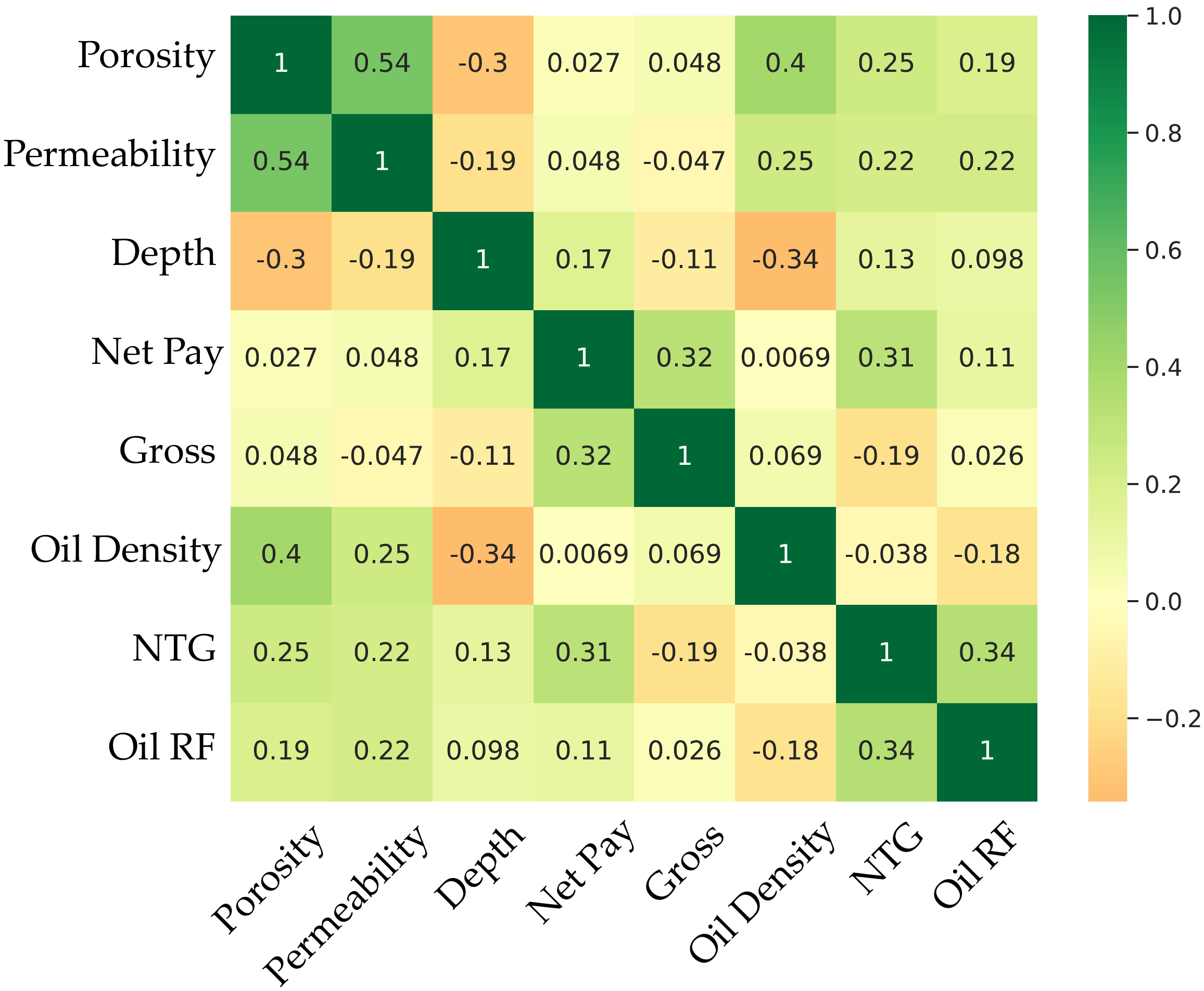}
\caption{Correlation matrix of continuous variables \label{correlation}}
\end{figure}

\ref{correlation} shows the correlation matrix of continuous variables. From this matrix, one can see the strong correlation between ``Porosity'' and ``Permeability'', ``Porosity'' and ``Oil Density''.

\begin{figure}
\includegraphics[width=1\textwidth]{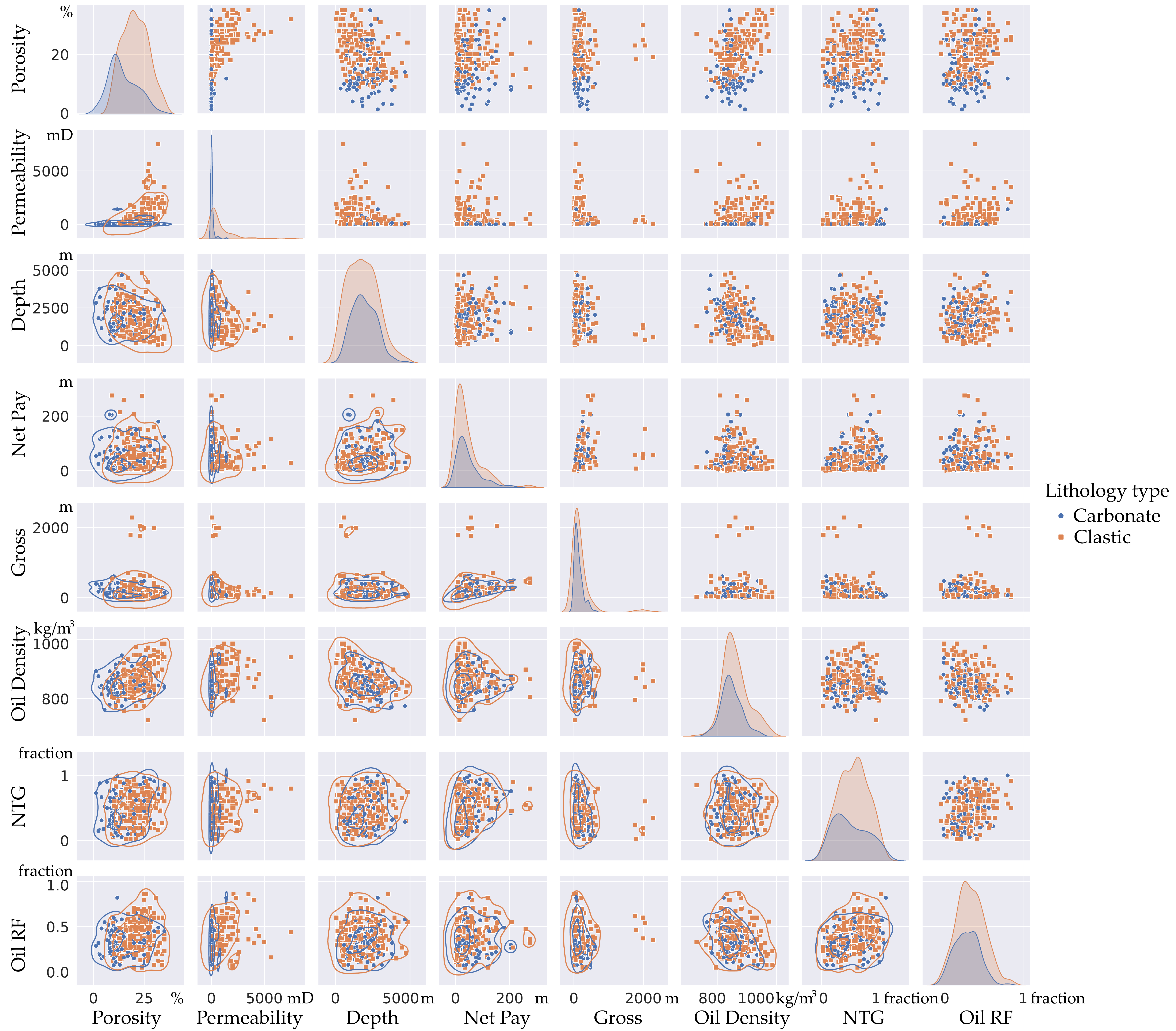}
\caption{Pair plots and histograms of continuous variables. The colour shows the distributions for different types of Lithology: Clastic is orange, Carbonate is blue.} \label{pairplot}
\end{figure}

Since we are working with a multivariate distribution, it is difficult to visualize. However, one can analyze one-dimensional parameter distributions as well as pairwise two-dimensional distributions of continuous parameters. Such distributions can be found in \ref{pairplot}.

Fig. \ref{pairplot} reveals several patterns, such as the more porous a clastic rock, the more heavy oil it contains (due to overlapping clastic regions in the corresponding block). Clastic rocks that lie in shallow depths are usually more porous due to less overburden pressure and may contain more heavy oil in comparison with rocks on deep depths due to oil degradation \cite{price1980crude}. This relationship may change for a different location, but the same pattern was revealed for Middle East carbonate reservoirs \cite{gomes2018benchmarking}. Reservoirs with lower NTG tend to have a lower RF. The poorer the reservoir quality, the lower the fraction of the reservoir that can produce oil. Porosity and permeability are closely related (fig. \ref{pairplot}) with higher porosities usually indicating higher permeability. The data show that clastic and carbonate reservoirs can have different porosity-permeability relationships.

\subsection{Data preprocessing}

\subsubsection{Discretization type}

\begin{figure}
\centering
\includegraphics[width=1\linewidth]{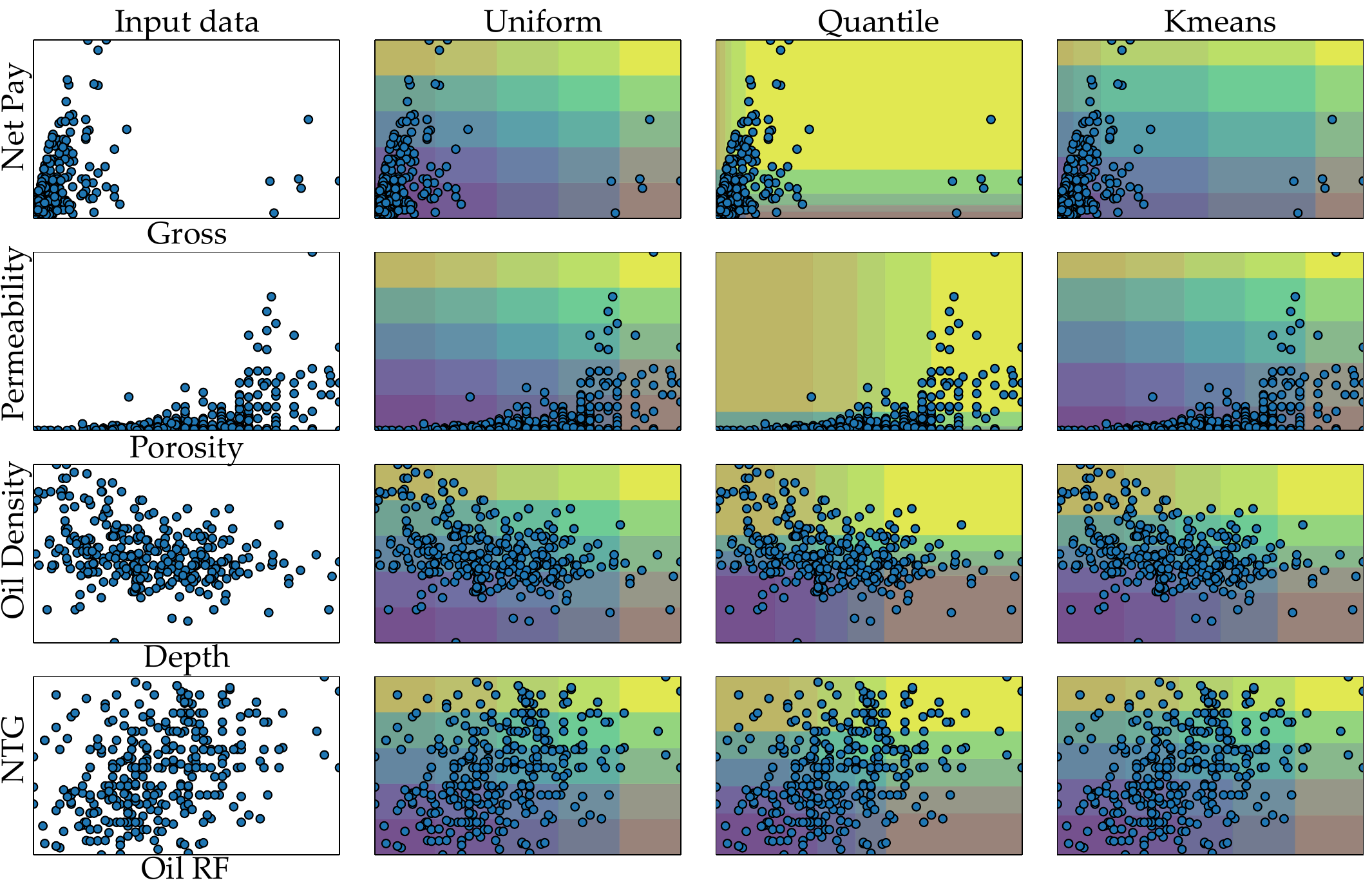}
\caption{Example of parameters discretization by different strategies for five bins. The first line shows the two-dimensional distribution of Net Pay from Gross, on the second the Permeability from Porosity, on the third - Oil Density from Depth and on the fourth NTG from Oil RF. For each  distribution, three discretization strategies are shown - Uniform, Quantile, and Kmeans. The intervals are shown with multi-coloured rectangles. Since the distribution is two-dimensional, discretization occurs along the X and Y axes. \label{disc}}
\end{figure}

The data must be discretized for the structure learning of BNs since the score functions are calculated based on discrete distributions. However, the discretization strategy can change the learning result and, consequently, the accuracy of modelling the parameters of the reservoirs. Three main discretization methods were considered: quantile, uniform and kmeans. In quantile discretization, the data is divided into intervals so that approximately the same amount of data fall into each interval, and the lengths of the intervals can be different. With uniform sampling, all data is divided into equal intervals. The kmeans algorithm for discretization implements the k-means clustering algorithm \cite{deeva2021oil}. The use of mathematical discretization methods is due to the need to use a single unified discretization method and the ability to discretize not only by the uniform distribution (because when we use geological information for discretization, this forms uniform distributions).  It is necessary to maintain a balance of complexity and accuracy when we select the number of intervals for discretization experimentally; it was determined that the number of intervals greater than 10 leads to a severe increase in the computation time. Thus, for the study, two numbers of intervals were taken - 5 and 10. An example of parameters discretization with five bins is shown in fig. \ref{disc}.

\subsubsection{Data normalization}
\label{normalization}
\begin{figure}
\centering
\includegraphics[width=1\linewidth]{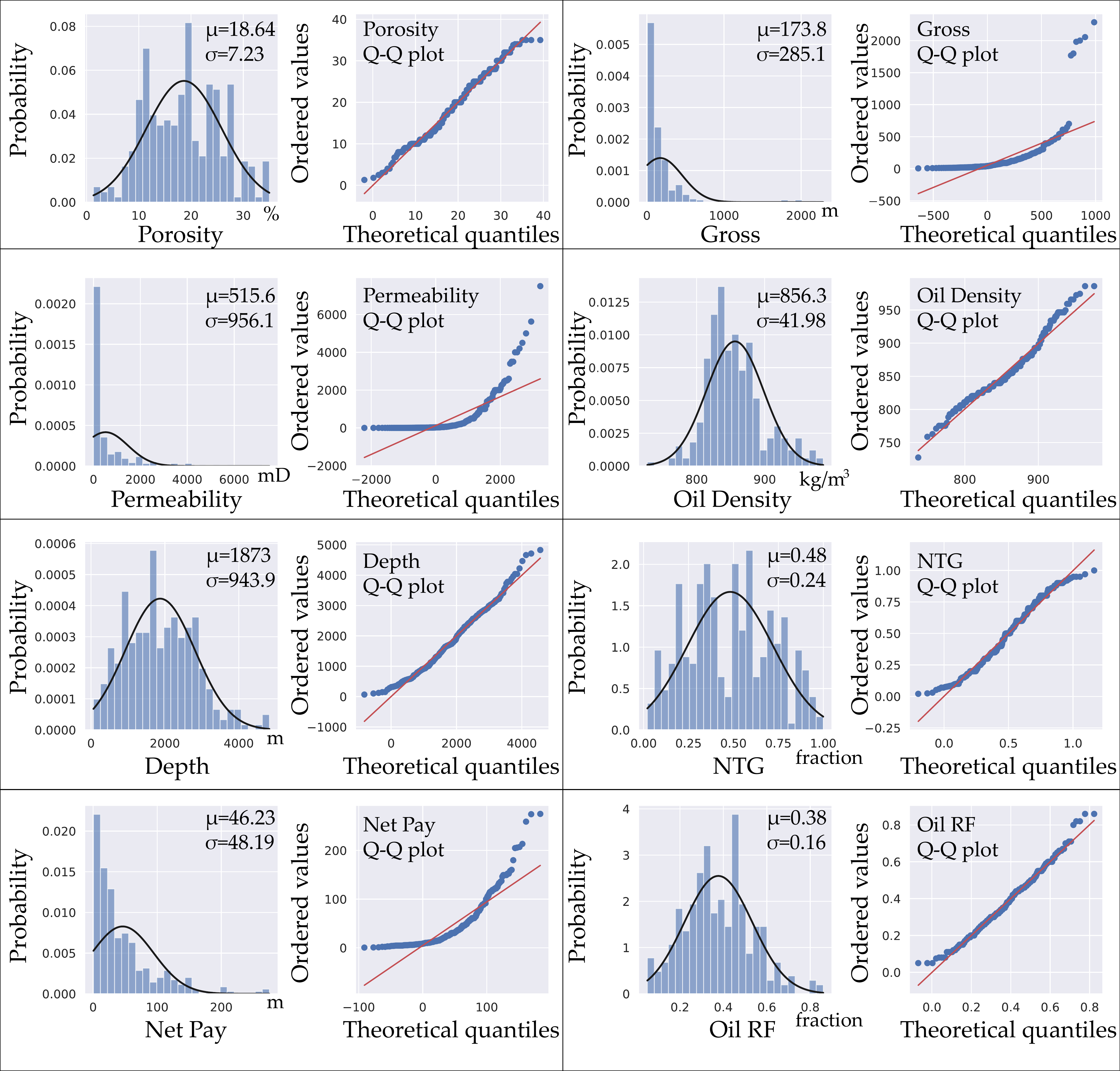}
\caption{Probability density functions with fitted normal distributions and Q-Q plots of continuous parameters: Porosity, Permeability, Depth, Net Pay, Gross, Oil Density, NTG and Oil RF.\label{qqlots}}
\end{figure}

The dataset includes eight continuous parameters: Porosity, Permeability, Depth, Net Pay, Gross, Oil Density, NTG and Oil RF. 
Fig. \ref{qqlots} shows the probability density function (PDF), fitted normal distributions and quantile-quantile (Q-Q) plots for each continuous parameter.

Permeability, Net Pay, and Gross deviate most from the normal distribution (fig. \ref{qqlots}), and so their logarithm values were considered for normalization. The question then arises: how much does the accuracy of the model change after normalization? 
The results of the influence of normalization of these parameters on the accuracy of the models can be found in the Section \ref{normalization_and_ordinal_results}.

\subsubsection{Ordinal categorical parameters}
\label{subsection_nominal_to_ordinal}
Categorical parameters can be both nominal and ordinal. However, the nominal values are not ordered and, consequently, are independent within the parameter distribution. Therefore, there is no single way of ordering values from the largest to the smallest for such parameters, and in average values have no meaning. Instead, there is just a list of different categories. Such parameters are, for example, Tectonic Regime, Structural Setting or Lithology. However, there are also ordinal parameters, such as the Period. Therefore, this parameter can be used for hierarchical filtering of resulting BNs, which would allow revealing robust dependencies on different geological ages.

\begin{figure}
\includegraphics[width=1\linewidth]{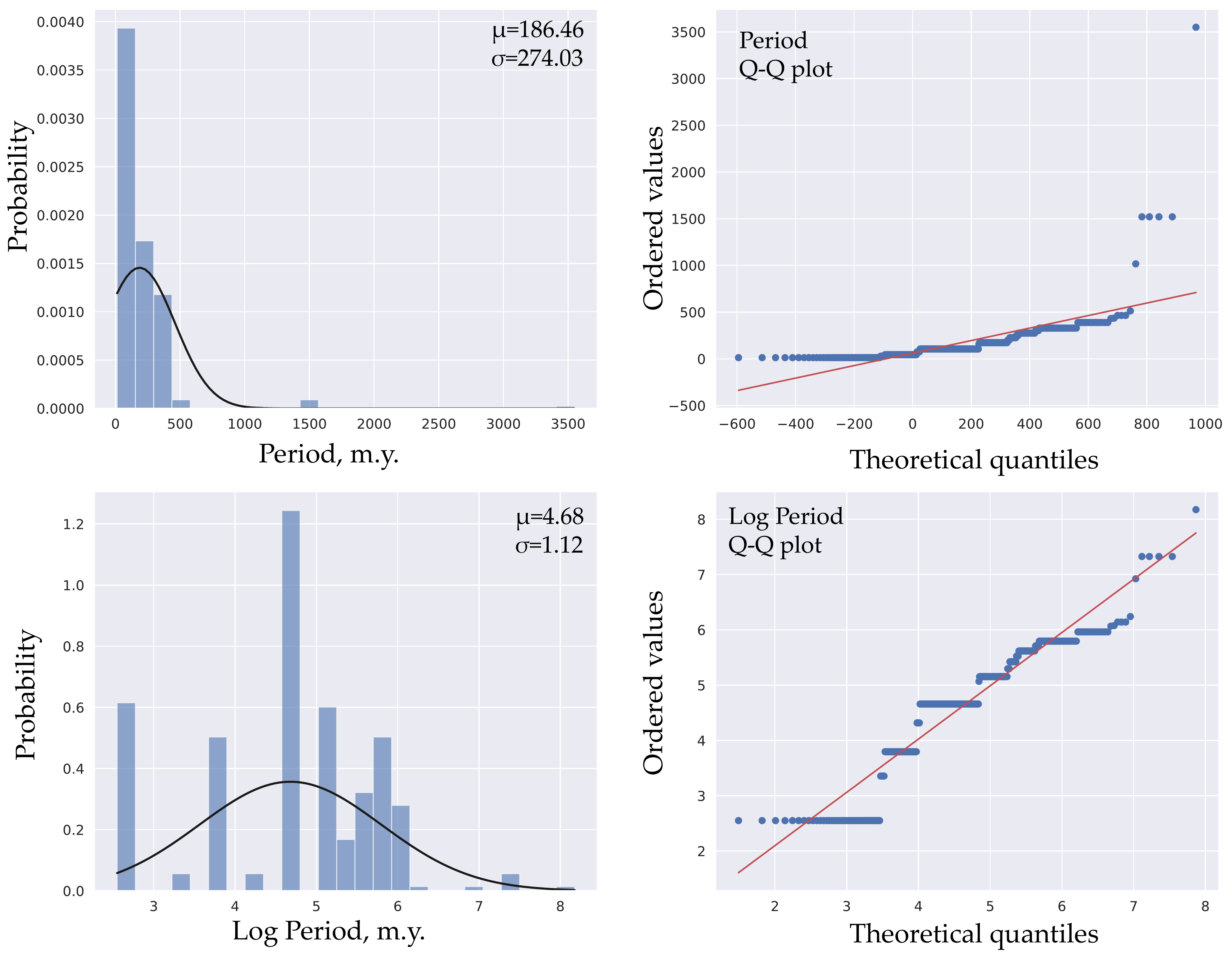}
\caption{Probability density functions with fitted normal distributions and Q-Q plots of continuous Period and logarithm of the Period.\label{period_plain}}
\end{figure}

The paper considers the hypothesis of increasing the accuracy by converting the Period parameter from an ordinal to a continuous one. Each period value has been replaced with the average age of the Period (the sum of the beginning and the end of the Period divided by two). The resulting distribution can be seen in fig. \ref{period_plain}. Since the distribution is very different from the normal, the distribution of the logarithm of the Periods was also considered.

The results of the influence of the transformation of ordinal categorical parameters to continuous on the accuracy of the models can be found in the Section \ref{normalization_and_ordinal_results}. 

\section{Bayesian Networks}

A Bayesian network is a directed probabilistic model whose structure is a directed acyclic graph \cite{koller2009probabilistic}. The BN reduces the dimension of the original multivariate distribution since its distribution is decomposed into a product of conditional distributions based on the network structure. The structure of a network can be a matter of expert judgment if it is known which nodes depend on each other. However, if there is no expert knowledge, the BNs can be learned from data.

\subsection{Structure learning}
\begin{figure}
\centering
\includegraphics[width=1\linewidth]{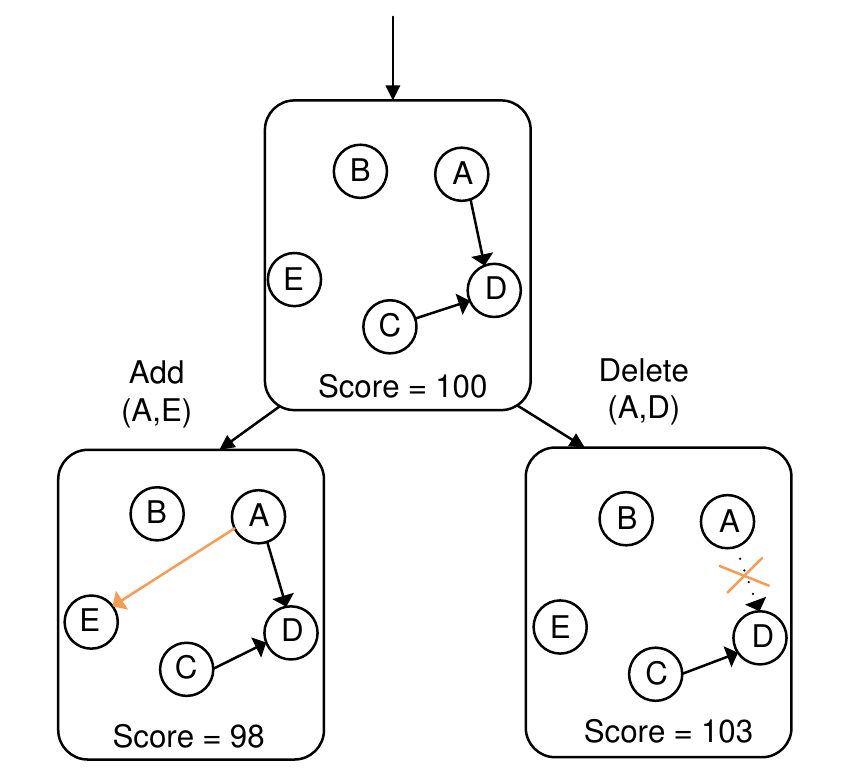}
\caption{An example of a greedy Hill-Climbing algorithm for structured learning.\label{hill_climb}}
\end{figure}

Among all algorithms for learning the structure of BNs, greedy heuristics are the most common algorithm \cite{chickering2002optimal}. In this article, we have used the Hill-Climbing algorithm \cite{gamez2011learning}. The algorithm starts by calculating the score function for an empty graph. Then, at each step, one action with an edge is checked (adding, deleting, or changing direction), the score function is calculated, and if the action leads to an increased score function, it is applied to the graph. The general scheme of the algorithm is shown in fig. \ref{hill_climb}. The score function is understood as a function that evaluates the quality of the BN structure in terms of maximizing the likelihood of the multivariate distribution that we are modelling. The kind of function can be different depending on the underlying assumptions about the distributions at the nodes \cite{carvalho2009scoring}.

If we consider the parameters of the reservoir as a multivariate probability distribution, then we need to find such a Bayesian network structure that would maximize the likelihood of the available data. That is why the structure of the network may differ from the physical dependencies between the parameters since it describes the configuration of such conditional distributions that would increase the likelihood of data given the structure of the network (P(D|G), where D - data, G - BN structure, P - probability). For better understanding the process of finding the structure of a Bayesian network, you can refer to pseudocode Alg.1.

\begin{algorithm}
\caption{Pseudocode of Bayesian Network structural learning algorithm with Hill-Climbing strategy. Here a function that evaluates the likelihood of data given a structure can be understood as a score.}
\label{BN_alg}
 \KwData{D, an initial empty DAG G}
 \KwResult{the DAG $G_{max}$ that maximises Score(G,D)}
 \SetKwData{SG}{$S_G$}
 \SetKwData{Smax}{$S_{max}$}
 \SetKwData{Gmax}{$G_{max}$}
 \SetKwData{GG}{$G^*$}
 \SetKwData{SGG}{$S_{G^{*}}$}
 \SetKwData{G}{$G$}
 \SetKwData{Gcurr}{$G_{current}$}
 \SetKwFunction{Score}{Score}
 
 \SG $\leftarrow$ \Score{G,D} \\
 \Smax $\leftarrow$ \SG \\
 \Gmax $\leftarrow$ \G \\
 \While{\Smax increases}{
 \For{all possible edges addition, deletion, reversal in \Gmax}
 {
   \SGG $\leftarrow$ \Score{\GG,D} \\
 \If {\SGG $>$ \Smax and \SGG $>$ \SG} 
 {\G$\leftarrow$\GG \\
 \SG $\leftarrow$ \SGG\\}
 }
 
  \If {\SG $>$ \Smax}
 {\Smax $\leftarrow$ \SG and \Gmax $\leftarrow$ \G

}

 }
\end{algorithm}

However, when using the Hill-Climbing algorithm, it is necessary to determine the score functions. Currently, the main score functions for training are information criteria (Bayesian information criterion (BIC), mutual information (MI)) \cite{scanagatta2019survey}, functions based on the Dirichlet distribution (K2) \cite{cooper1992bayesian}. There are also variations of these functions for mixed distributions ($MI_{mixed}$, $BIC_{mixed}$) \cite{bubnova2021}.

\subsection{Parameter learning}
\begin{figure}
\centering
\includegraphics[width=1\linewidth]{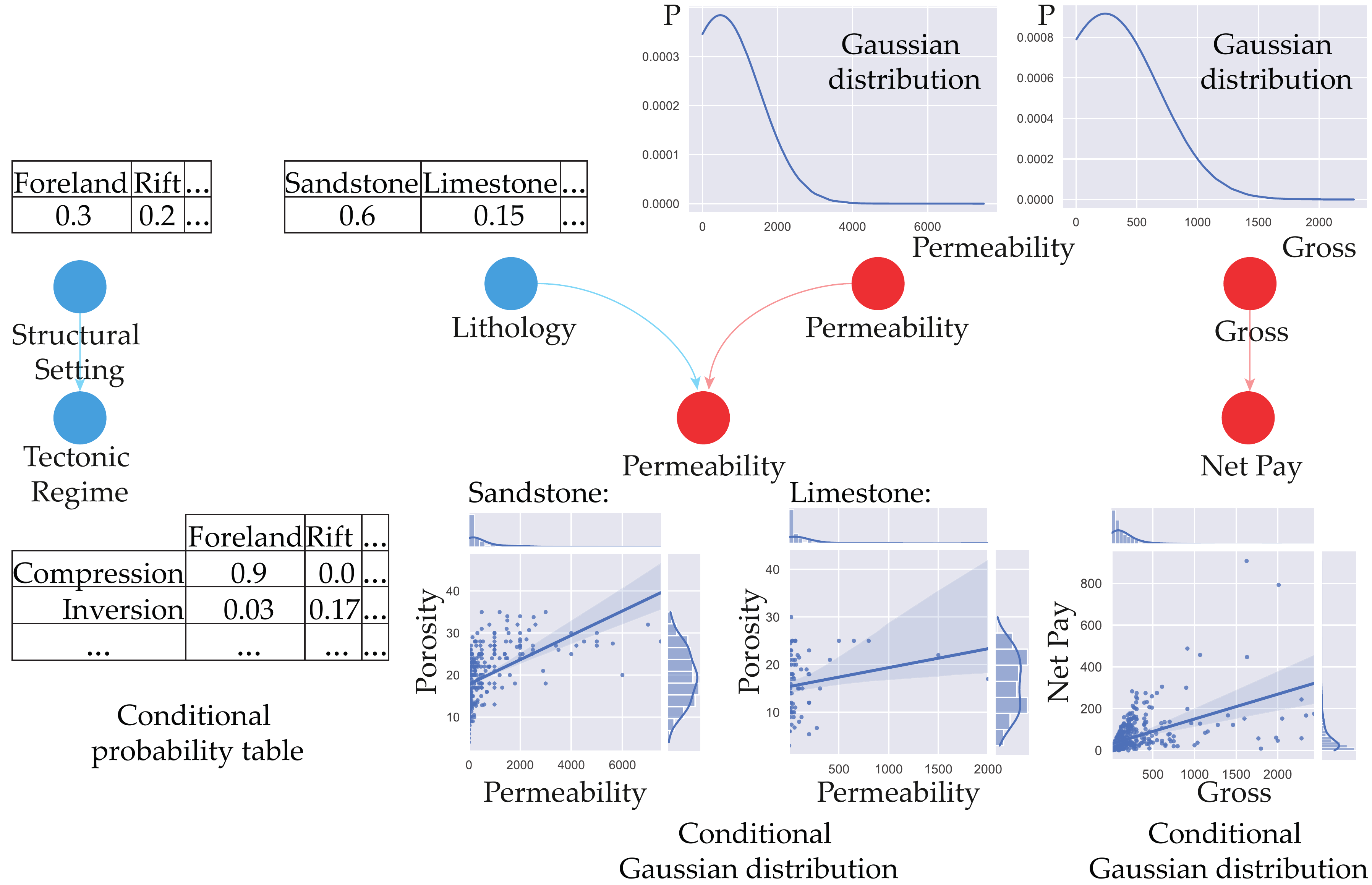}
\caption{An example of parameter learning at the nodes of a BN. It contains such cases (from left to right): a discrete node with a discrete parent (left), a continuous node with discrete and continuous parents (centre), a continuous node with a continuous parent (right).\label{param_learn}}
\end{figure}

After learning the BN structure, it is necessary to learn the distribution parameters in the nodes. In this study, the likelihood maximization method \cite{koller2009probabilistic} is used to train the parameters. Since the data contain both discrete and continuous values, we should choose a learning strategy that handles mixed data. Often, in the case of continuous data, they are discretized, but the information is lost, and modelling accuracy decreases. Therefore, a method was used that makes it possible to use data on the initial distribution of continuous nodes. This approach uses conditional probability tables for discrete nodes, Gaussian distributions for continuous nodes, and conditional Gaussian distributions for mixed nodes \cite{bottcher2001learning}. As you can see in fig. \ref{param_learn}, if a node is discrete and / or has discrete parents, then its distribution is described by a probability table. If a node is continuous and has continuous parents, then this dependence is described by linear regression, and if a discrete parent is present, then regressions are found for each combination of a discrete parent.

\section{Search for similar reservoirs}

\subsection{Search methods reservoirs analogues}

When analysing oil and gas reservoir data, we are sometimes interested in one fixed target reservoir. A standard method of investigating this target is to examine a subsample of neighbouring or similar objects to the target. This subsample is usually called a set of analogues. Once identified, it is possible to build a BN on this set, perform estimation, check for anomalies, or predict missing values. The main issue is only to determine a suitable measure of proximity or distance to the target reservoir. The main difficulties in choosing a metric are described in our previous work \cite{deeva2021oil}, and here we present the best metrics for use in our experiments.

Let $u$ and $t$ be compared objects. Consider standard similarity measures for a fixed, variable $j$. For quantitative variables, we will use the following:
\begin{equation}
S_j(u,t)=\frac{|u_j-t_j|}{1-\min\limits_x(x_j)/\max\limits_x(x_j)}.
\end{equation}

This normalisation is part of the Gower library in Python and differs from the classical version.
For categorical variables, we check for matching values:
\begin{equation} 
S_j(u,t)=
\begin{cases}
1, &u_j=t_j \\ 
0, &u_j \neq t_j\\
\end{cases}.
\end{equation}

From $S_j(u,t)$ we can get the overall Gower similarity coefficient:
\begin{equation}
S(u,t)=\frac{\sum\limits_{j=1}^p w_{jut}\cdot S_j(u,t)}{ \sum\limits_{j=1}^p w_{jut}}.
\end{equation}

Here $w_{jut}$ is the weight on the variable $j$ for $u$ and $t$ and unless otherwise stated:
\begin{equation}
w_{jut} =
\begin{cases}
1, &\text{if reservoirs "u" and "t" are being compared} \\ 
0, &\text{if reservoirs "u" and "t" are not being compared}
\end{cases}.
\end{equation}

And when it comes to distance, consider the following:
\begin{equation}
dist_G(u,t)=1-S(u,t).
\end{equation}

In the case where we only work with categorical variables, this distance reduces to the Hamming distance:
\begin{equation}
dist_H\left(u,t\right)=p-\sum_{j=1}^{p}{S_j\left(u,t\right)}.
\end{equation}

Then:
\begin{equation}
dist_G\left(u,t\right)=1-\frac{\sum_{j=1}^{p}{S_j\left(u,t\right)}}{p}=\frac{1}{p}dist_H\left(u,t\right).
\end{equation}

There are also measures that differ from the Hamming distance or the Gower coefficient. For example, in the task of ranking search engine results, measure cosine distance for vectors $u$ and $t$:
\begin{equation}
1-\frac{\sum_{i}{w_i u_i t_i}}{\sqrt{\sum_{k}{w_k u_k^2}}\sqrt{\sum_{k}{w_kt_k^2}}}.
\end{equation}

An unweighted version of this distance:
\begin{equation}
1-\frac{\sum_{i}{u_it_i}}{\sqrt{\sum_{k} u_k^2}\sqrt{\sum_{k} t_k^2}}=1-cos\left(u,t\right).
\end{equation}

\begin{figure}
\begin{subfigure}[b]{.30\textwidth}
  \centering
  \includegraphics[trim=0 0 0 0, clip, width=1.0\linewidth]{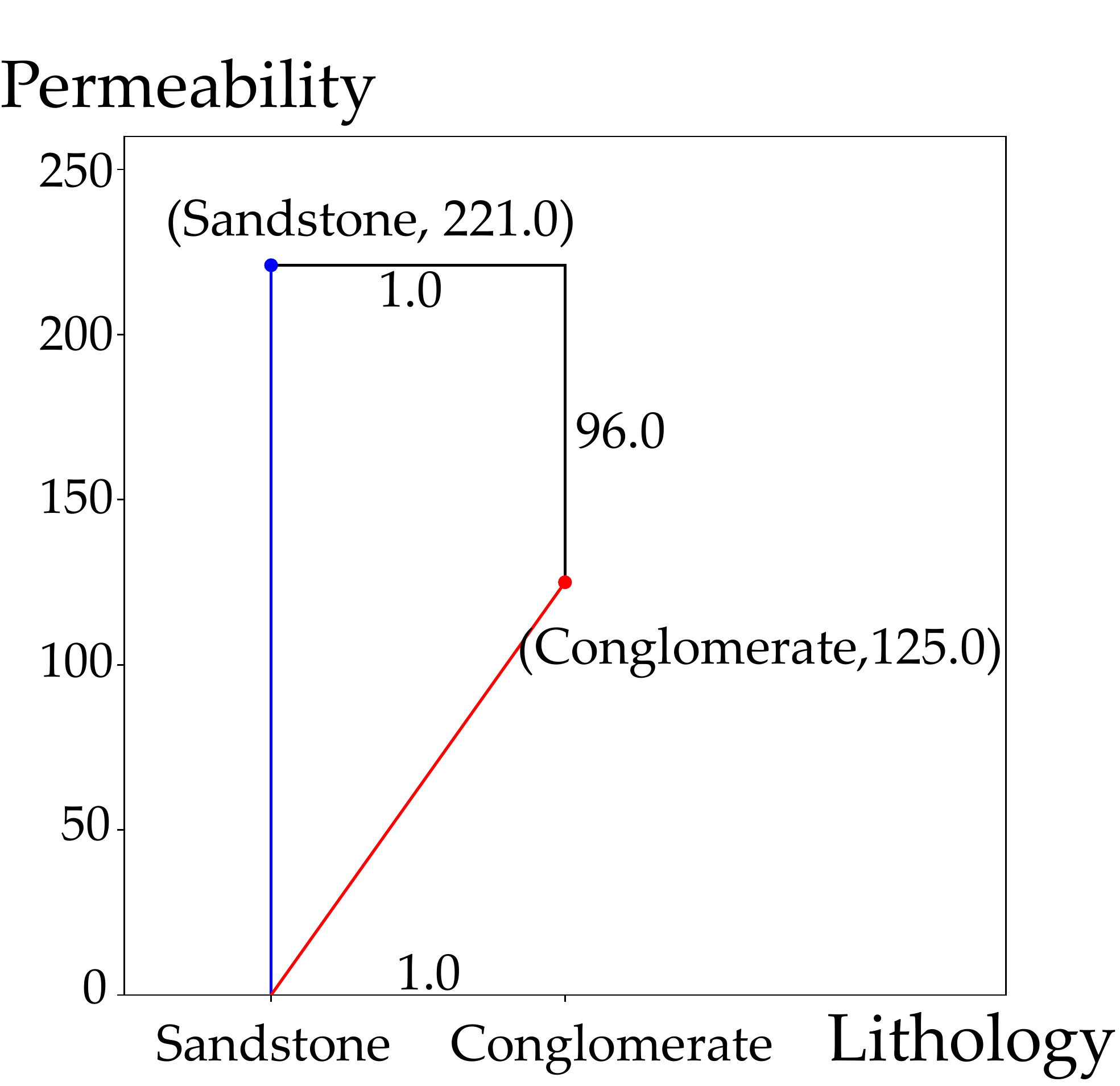}\\a)
\end{subfigure}
\begin{subfigure}[b]{.30\textwidth}
  \centering
  \includegraphics[trim=0 0 0 0, clip, width=1.0\linewidth]{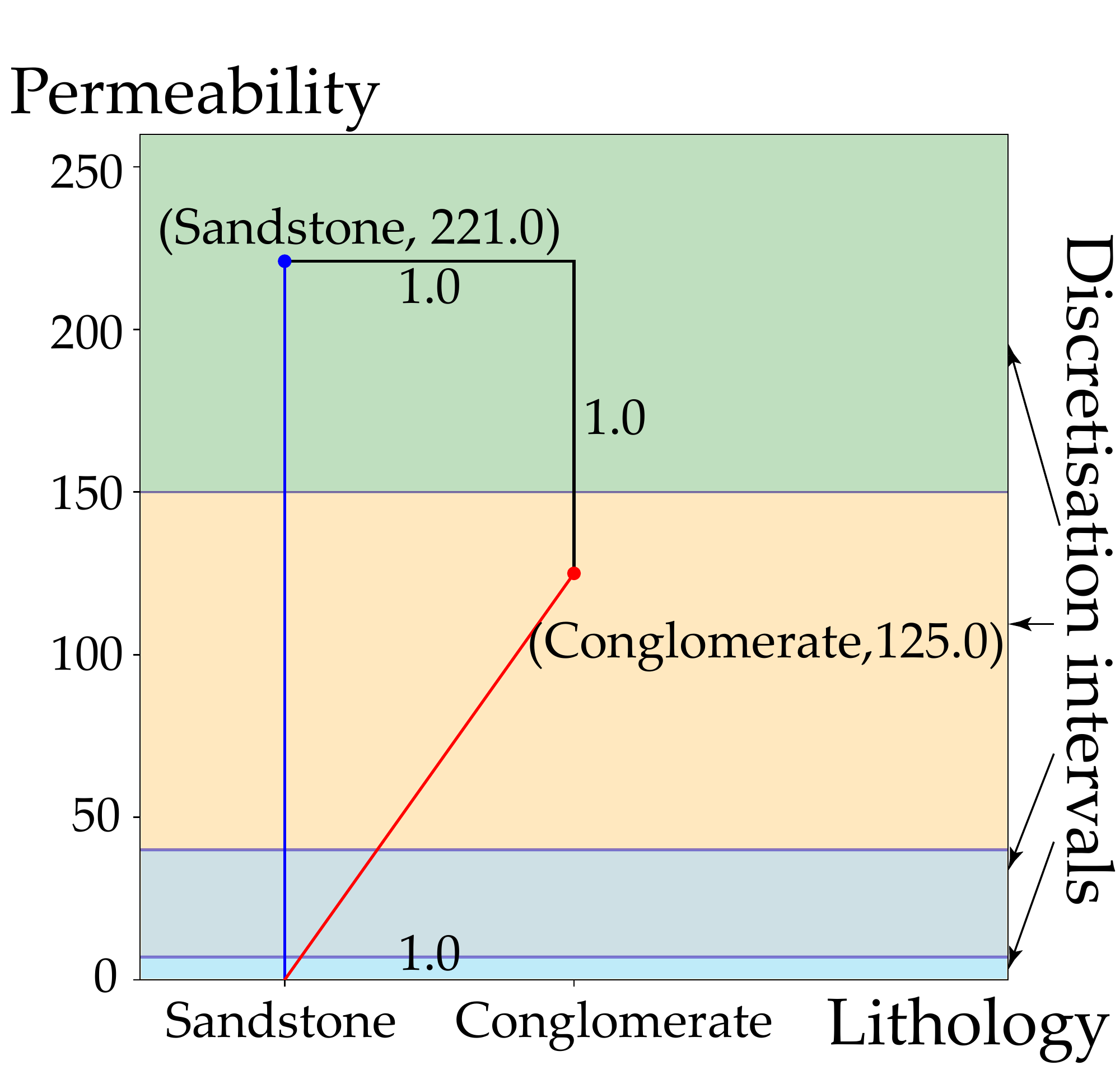}\\b)
\end{subfigure}
\begin{subfigure}[b]{.30\textwidth}
  \centering
  \includegraphics[trim=0 0 0 0, clip, width=1.0\linewidth]{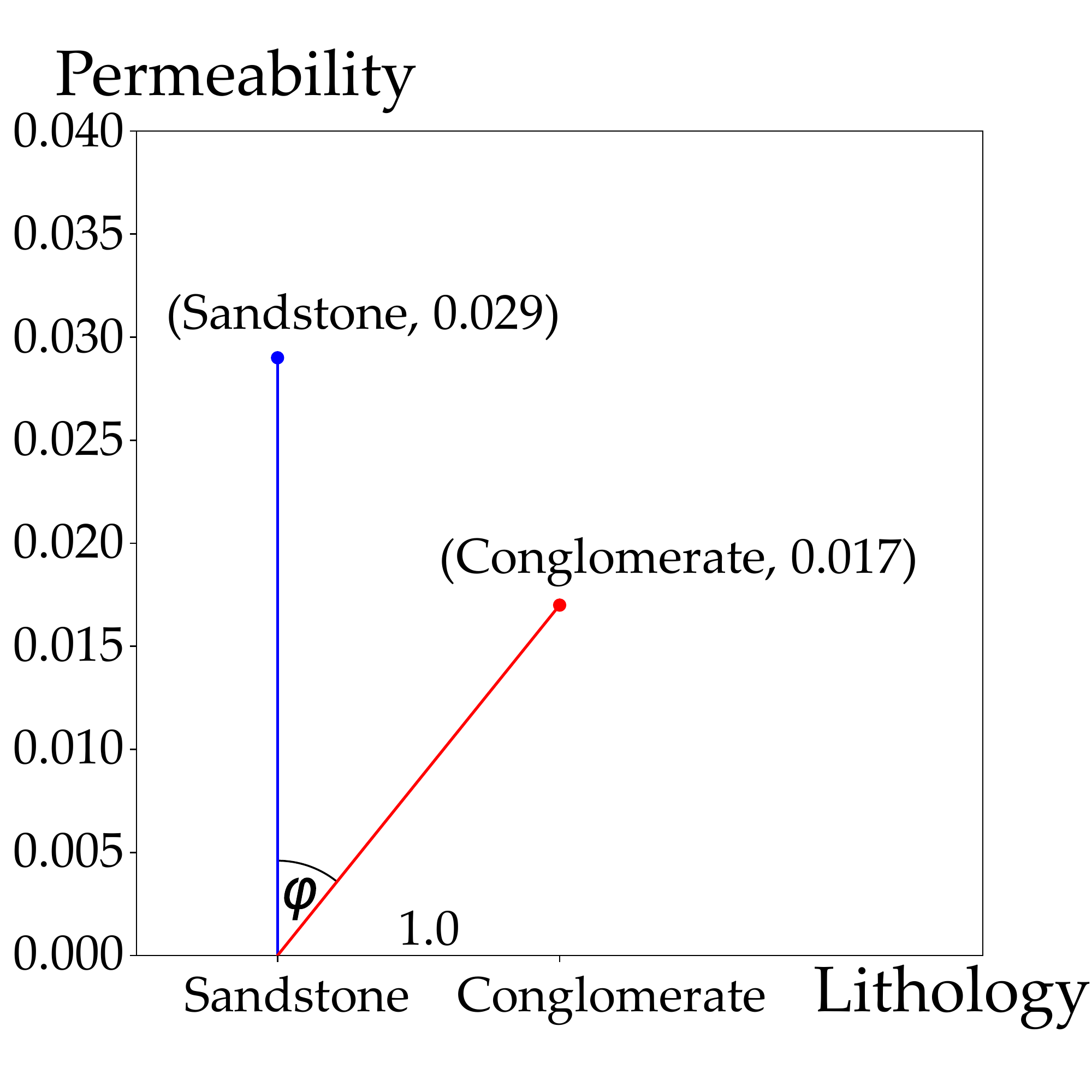}\\c)
\end{subfigure}
\caption{Example of distance calculation from \{Lithology: SANDSTONE, Permeability: 221.0\} to \{Lithology: CONGLOMERATE, Permeability: 125.0\}. The figures show data after normalization for Gower (a), Hamming (b) and cosine (c) distances. Normalization for (a) and (b) did not result in a change because the minimum permeability value on these data is 0. The cosine distance (c) in this case is $1-cos(\varphi)$.
\label{visual_dist}}
\end{figure}

This variant has a geometric interpretation through the angle between vectors $u$ and $t$. The distance is 0 when the vectors are congruent and 1 when orthogonal.
Applying this distance requires a prior transformation of values which is described below. For a categorical variable, the value is assumed to be one at the target—moreover, 0 or 1 on the object being compared, depending on whether the categories match.
For quantitative variables, the values undergo the following transformation:
\begin{equation}
u'_j = \frac{u_j-\min_x (x_j)}{range(j)}
\end{equation}
where $ range(j)=\max_x (x_j)-\min_x (x_j)$. \ref{visual_dist} shows examples of all the distances mentioned with the corresponding normalizations.

All of the above distance measures can be used with weighting factors. By weighting factors, we mean the coefficients that increase or decrease the penalty for a certain parameter. In general, it is quite difficult to determine which parameters are important and which are not in the context of the problem, and to assign proportional weights to them. Such weights can be obtained by two approaches: 1) with the help of specialized knowledge of domain experts or 2) with the help of statistical analysis and optimization methods. In our previous work \cite{deeva2021oil}, we tried to solve this problem by introducing an additional regularization for continuous parameters using estimation weights. For this purpose, we relied on the analysis of, for example, the Gower distance, for which the penalties for discrete parameters are on average higher than for continuous parameters. Note that in this study all parameters are taken with a weight equal to one, but our methods allow us to take expert knowledge into account at this stage as well.

\subsection{Clustering based on BNs}
\label{bn_clustering}
One way to represent analogues is based on the clustering of reservoirs. In this formulation, we consider clusters as a set of similar reservoirs. However, in our case, the difficulty lies in the fact that the data are not only continuous values that can be clustered, for example, through Euclidean distance, but also discrete values. Existing solutions involve, for example, reducing the initial space of heterogeneous data using principal component analysis (PCA) \cite{martin2013new}, and then clustering such data. However, the disadvantage of this approach is that by decreasing the dimension, we lose much information.

\begin{figure}
\centering
\includegraphics[width=1\linewidth]{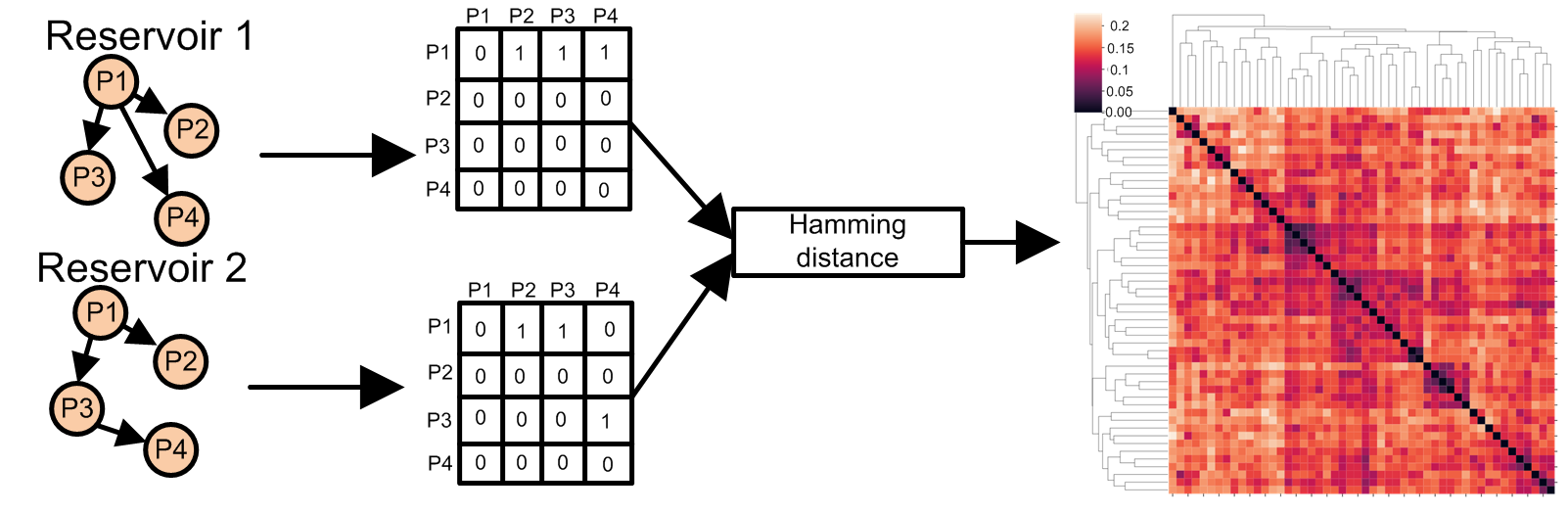}
\caption{Pipeline clustering algorithm based on BNs structures. It contains such steps: construct BNs on subsamples, get adjacency matrices of BNs, calculate Hamming distance between matrices and cluster the final matrix with distances. \label{bn_cluster}}
\end{figure}

The use of BNs allows us to consider the relationships between parameters during clustering and thus cluster data in the space of joint parameter distributions. BNs clustering algorithm is shown in \ref{bn_cluster} and includes the following steps:
\begin{enumerate}
    \item The reservoir is taken from the dataset, and the N nearest according to one of the distance metrics is searched;
    \item Then, on the obtained subsample, a BN is built, and its structure is preserved;
    \item The steps are repeated for each reservoir;
    \item Then, the network structures are hierarchical clustered based on Hamming distance.
\end{enumerate}

A Random Forest classifier was trained to predict clusters by input reservoir parameter. This classifier model was chosen as we do not have much data, and classes may be imbalanced.

\subsection{Bayesian networks on filtered data}
\begin{figure}
\begin{subfigure}[b]{.55\textwidth}
  \centering
  \includegraphics[width=0.95\linewidth]{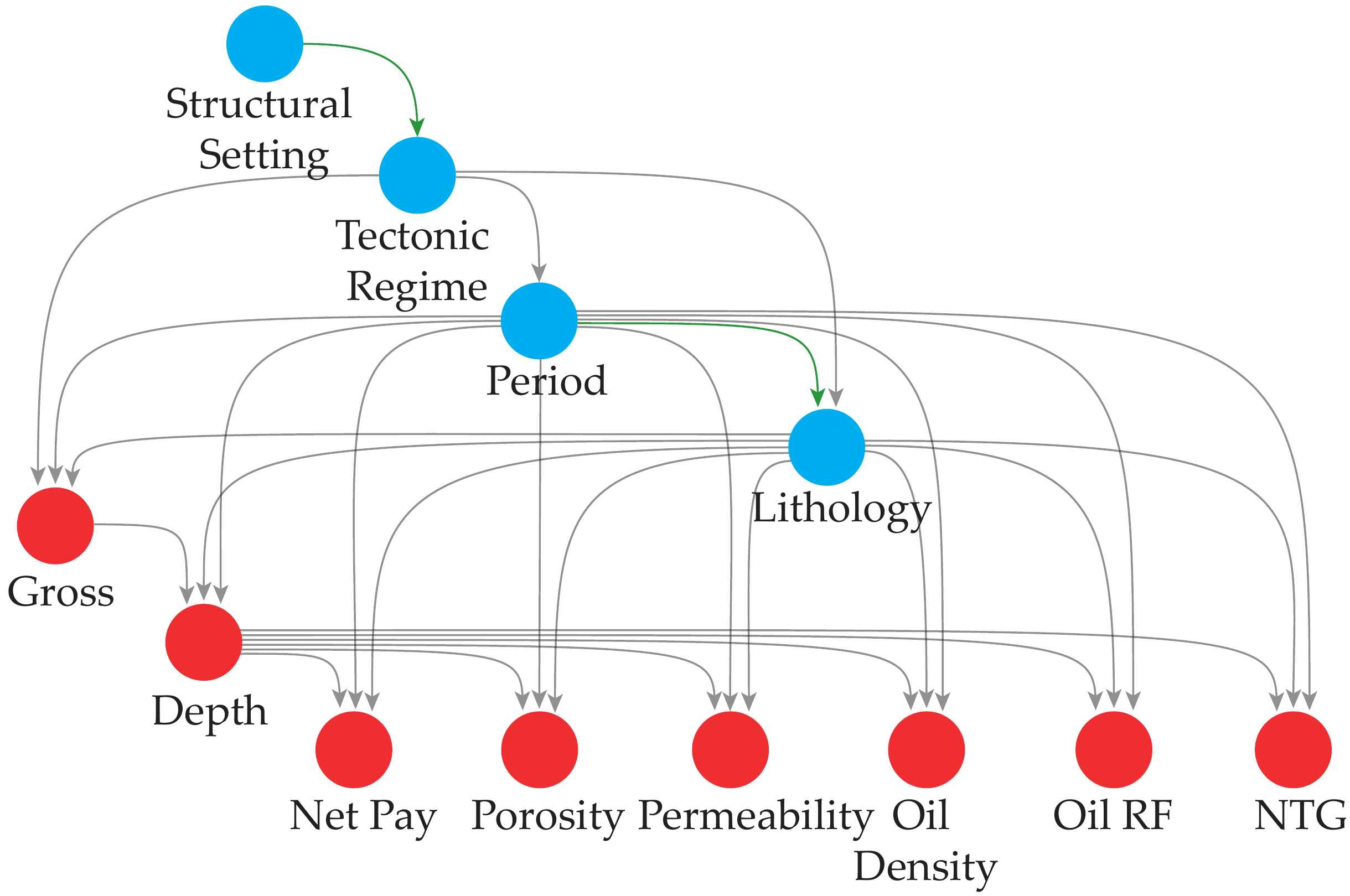}\\a)
\end{subfigure}%
\begin{subfigure}[b]{.55\textwidth}
  \centering
  \includegraphics[width=0.95\linewidth]{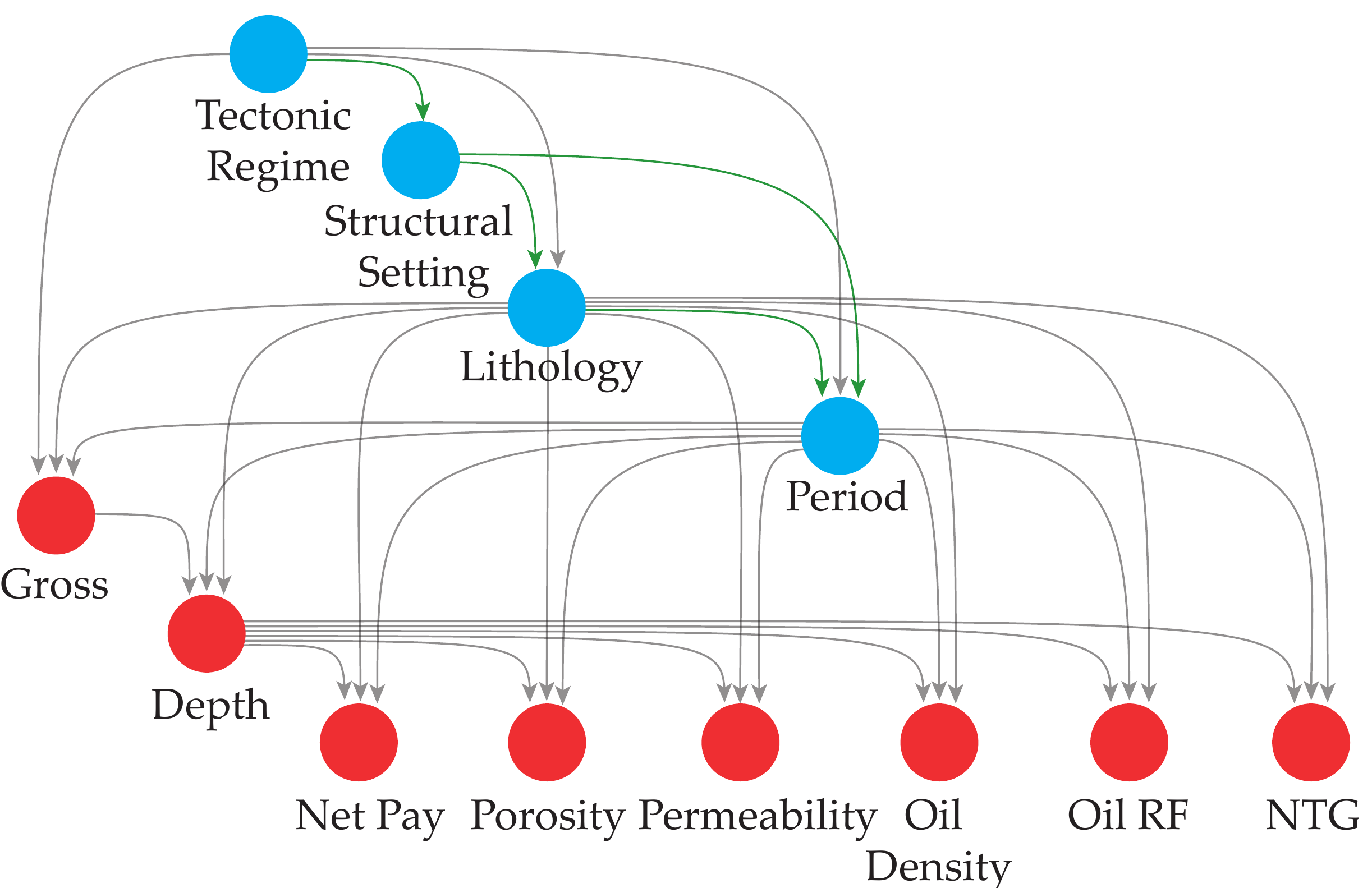}\\b)
\end{subfigure}
\caption{Example of two BNs built on filtered data by the Structural Setting parameter. These networks build only on reservoirs whose Structural Setting is Salt (a) or Inversion (b). The common stable edges for this group are highlighted in grey. The different edges are highlighted in green, and they will be summed when calculating the distance for this pair of networks. 
\label{salt_rift}}
\end{figure}

In this subsection, we explore the idea of extracting stable relationships between variables by constructing BNs on filtered data in which some variables have fixed values (for example, separate BN with Period=Triassic, or Lithology=Sandstone). We can distinguish identical and different edges for two networks based on filtered data. Identical or common edges indicate a stable relationship between pairs of variables for these two groups of data. The different or unstable edges are summed, and the resulting metric, the Hamming distance for graphs can be used further for hierarchical clustering into data groups for which BNs have more stable edges. fig. \ref{salt_rift} shows an example of BNs for data with Salt and Inversion values for the Structural Setting. 

\begin{figure}
\includegraphics[width=0.4\linewidth]{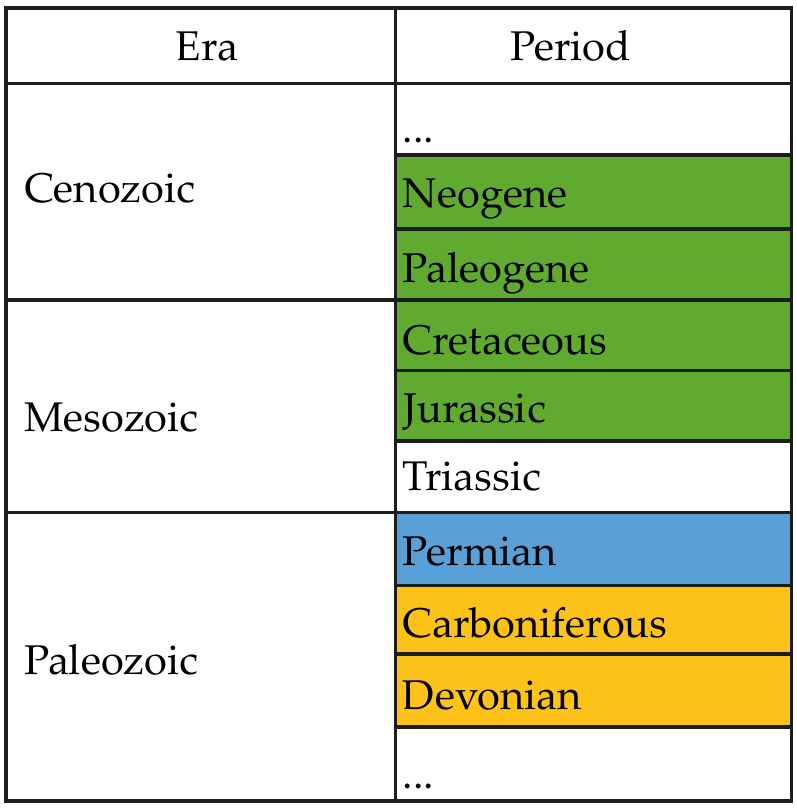}
\includegraphics[width=0.6\linewidth]{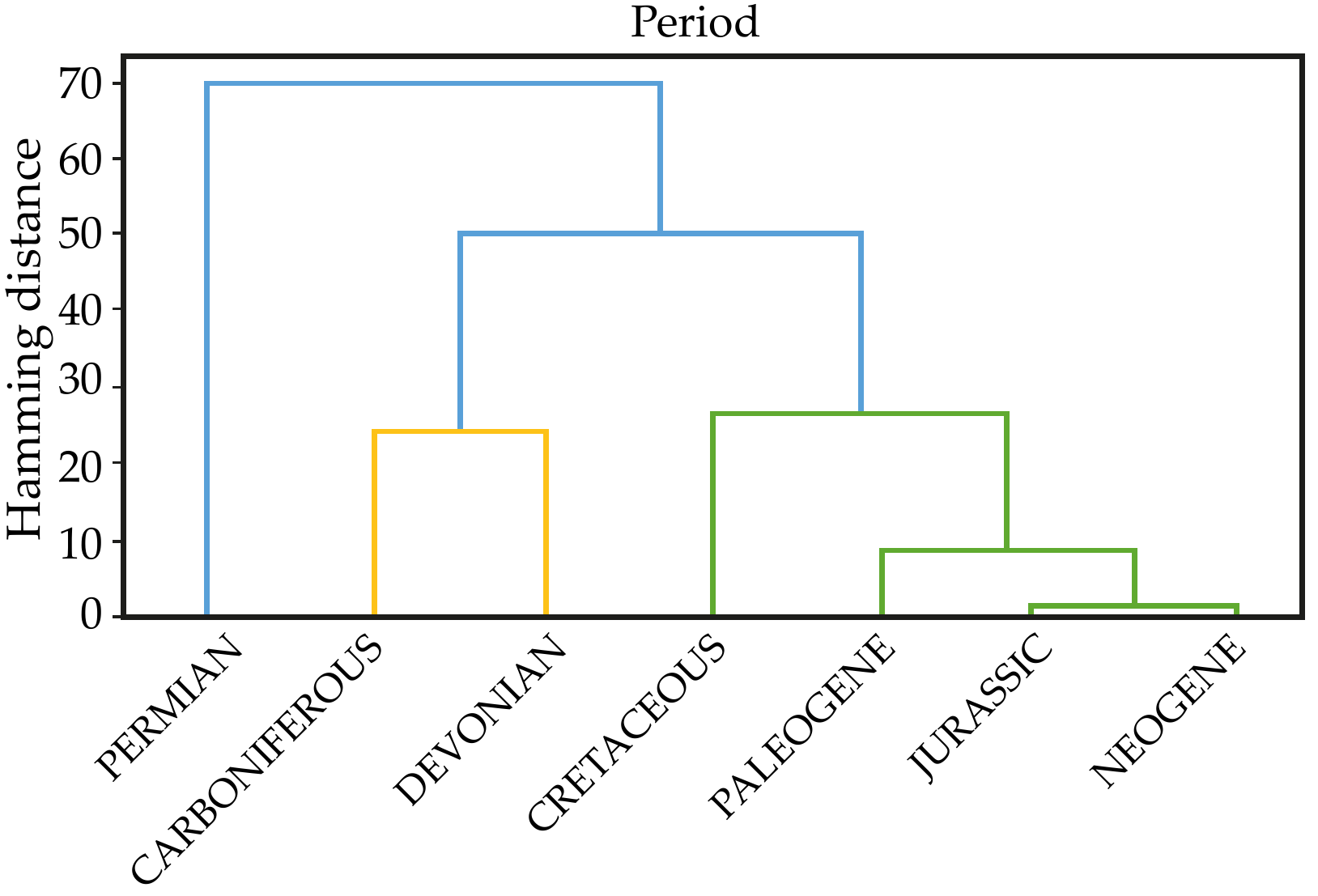}
\caption{Part of stratigraphic scale and dendrogram for period with clusters highlighted in color. \label{period}}
\end{figure}

The hierarchical clustering  result (fig. \ref{period}) shows that the networks structured on datasets filtered by  geologic periods are close together. For instance, Neogene, Paleogene, Cretaceous and Jurassic periods end up in the same cluster. At the same time, they are located one after the other on the stratigraphic scale.  Because clusters obtained by filtered BN are mimicking the natural distribution of parameter these results may suggest that such networks can increase the accuracy of predictions.

In the general case, such clustering can be used to solve problems of modelling the distribution or imputation of values. Modelling was done in the standard way for BNs, but the number of samples for each cluster was proportional to the size of the filtered data on the values corresponding to that cluster. Predicting the most likely target reservoir parameters of interest begins by identifying the cluster to which the reservoir belongs. The BN corresponding to that cluster can then calculate the most likely values for the missing parameters. In fact, the domain expert can filter not only by a single value, but by different sets of parameter values. For example, an expert can build networks on data, in which on one side the Structural Settings parameter will take Inversion, Wrench, Foreland values. And on the other side take Rift and Salt values. Thus, the domain expert can perform a very deep cause-effect analysis in various subsamples of the dataset.

\section{Experimental results}

\subsection{General Bayesian Network Experiments}
\subsubsection{Determining the best algorithm for Bayesian Networks}

In the previous sections, we established that different score functions and data discretization methods could be used to train the BN structure. In order to choose the optimal combination of function and discretization methods, the quality of modelling parameters for each network was calculated.
The quality of modelling by BNs is assessed by calculating the accuracy of missing values prediction for the parameters of oil and gas reservoirs. Since there are just a few hundred of samples (although each represents a reservoir), all the experiments use leave-one-out (LOO) validation. During imputation, the parameter value is deleted, the rest of the parameters are initialized to known values, and the unknown parameter is sampled. If the parameter is discrete, then it is imputed by the most frequent category in the sample; if it is continuous, it is imputed by the average value in the sample. For discrete parameters, the imputation accuracy is evaluated; for continuous parameters, normalized root mean square error (NRMSE), which is normalized by the parameter range, is evaluated.

\begin{figure}
\centering
    \includegraphics[width=0.9\linewidth]{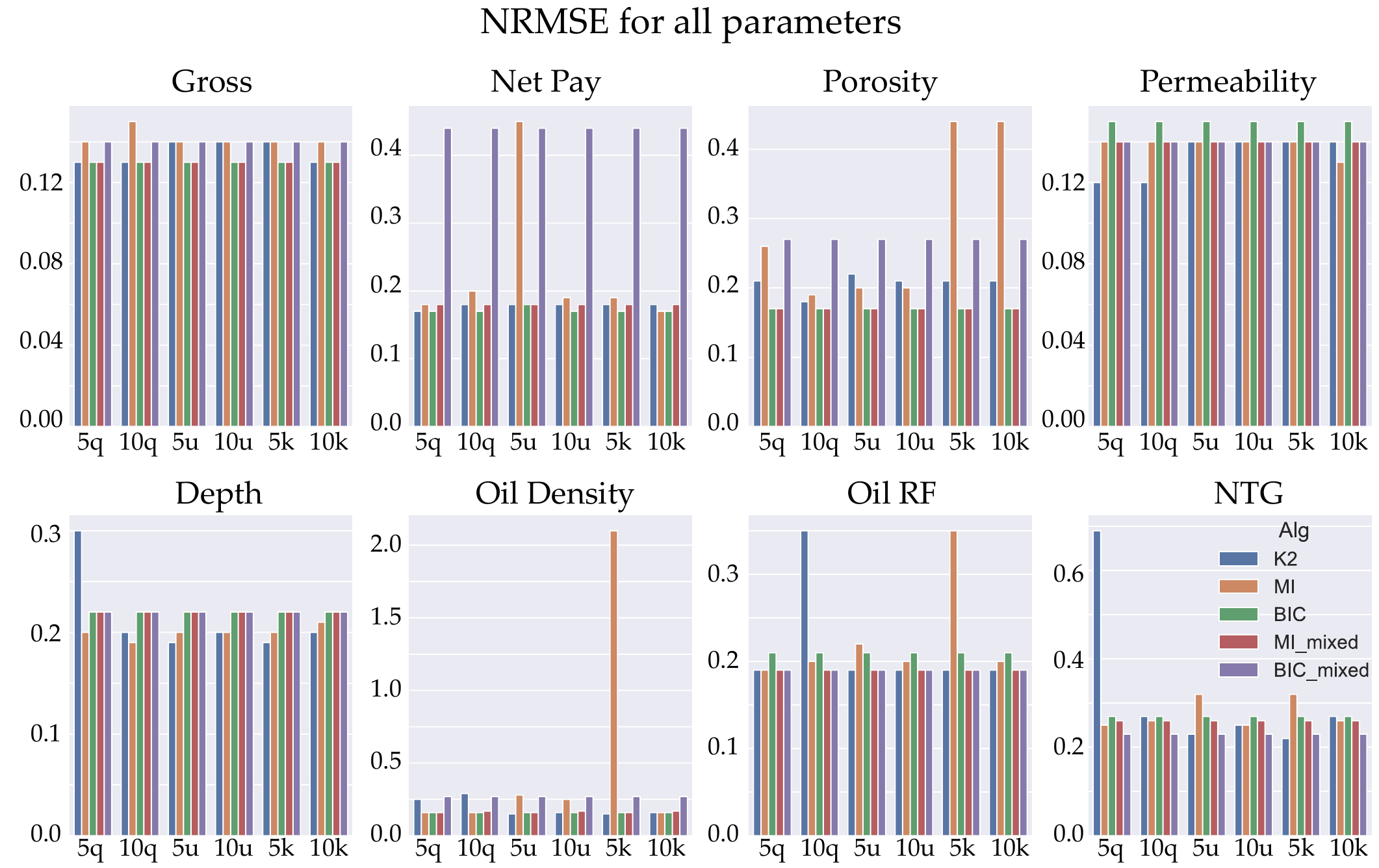}
    \caption{Prediction results (missing values imputation) for each continuous parameter. Y-axis records the NRMSE. The lower the value of NRMSE, the more accurate. The X-axis records the type of discretization (q - quantile, u - uniform, k - kmeans. The number indicates the number of bins). The colour represents the score function (blue - K2, yellow - MI, green - BIC, red - mixed MI, purple - mixed BIC). \label{rmse_all}}
\end{figure}

\begin{figure}
\centering
    \includegraphics[width=0.9\linewidth]{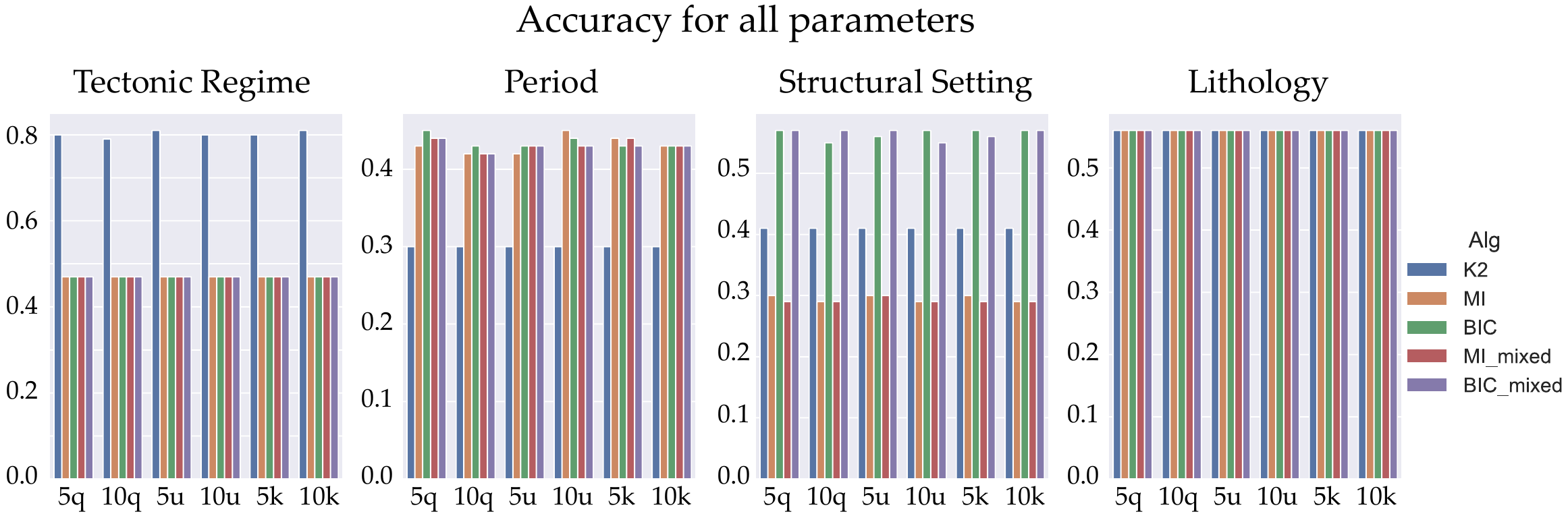}
    \caption{Prediction results for each discrete parameter.  The Y-axis records the accuracy of missing values prediction.  The higher the value on the y-axes, the more accurate. The X-axis records the type of discretization (q - quantile, u - uniform, k - kmeans, the number indicates the number of bins), color means score function (blue - K2, yellow - MI, green - BIC, red - mixed MI, purple - mixed BIC). \label{acc_all}}
\end{figure}

Fig. \ref{rmse_all} and fig. \ref{acc_all} show the experimental results for each parameter and for each combination of score function and discretization type. Accuracy and NRMSE for different parameters vary on average, although there are also individual spikes within specific parameters.

Since the scatter of results for different algorithms and parameters is quite large, some averaging of the results for all parameters is necessary to choose the most advantageous combination of the score function and discretization type. To do this, we calculated the average error value for continuous parameters and the average accuracy for discrete parameters for each combination of the score function and the type of discretization. Table \ref{mean_rmse_acc} shows the mean error and accuracy results for various combinations of the structure learning algorithms. Although the best NRMSE on continuous parameters is achieved using kmeans (5 bins) and K2, this algorithm also shows comparable accuracy on discrete parameters. This combination will thus be used for further experiments for structural learning of the network.

\begin{table}
\caption{Mean error for continuous parameters and mean accuracy for discrete parameters for each combination of the structured learning score (horizontally) and discretization type (vertically). q - quantile, u - uniform, k - kmeans, the number indicates the number of bins. \label{mean_rmse_acc}}
\centering
\begin{tabular}{|c|c|c|c|c|c|}
\hline
               & \textbf{K2}     & \textbf{BIC} & \textbf{MI} & \textbf{BIC\_mixed} & \textbf{MI\_mixed} \\ \hline
\multicolumn{6}{|c|}{\textbf{Mean NRMSE}}                                                                \\ \hline
\textbf{5k}  & \textbf{0.1775} & 0.185        & 0.485       & 0.2375              & 0.1813             \\ \hline
\textbf{10k} & 0.185           & 0.185        & 0.214       & 0.2375              & 0.183              \\ \hline
\textbf{5q}  & 0.2575          & 0.185        & 0.19        & 0.2375              & 0.181              \\ \hline
\textbf{10q} & 0.215           & 0.185        & 0.186       & 0.2375              & 0.183              \\ \hline
\textbf{5u}  & 0.18            & 0.186        & 0.244       & 0.2375              & 0.181              \\ \hline
\textbf{10u} & 0.184           & 0.185        & 0.196       & 0.2375              & 0.183              \\ \hline
\multicolumn{6}{|c|}{\textbf{Mean accuracy}}                                                             \\ \hline
\textbf{5k}  & 0.5175          & 0.5075       & 0.4425      & 0.5050              & 0.44               \\ \hline
\textbf{10k} & \textbf{0.52}   & 0.5075       & 0.4375      & 0.5075              & 0.4375             \\ \hline
\textbf{5q}  & 0.5175          & 0.5125       & 0.44        & 0.51                & 0.44               \\ \hline
\textbf{10q} & 0.5150          & 0.5025       & 0.435       & 0.5050              & 0.435              \\ \hline
\textbf{5u}  & \textbf{0.52}   & 0.5050       & 0.4375      & 0.5075              & 0.44               \\ \hline
\textbf{10u} & 0.5175          & 0.51         & 0.4425      & 0.5025              & 0.4375             \\ \hline
\end{tabular}
\end{table}

\subsubsection{Influence of normalization and transformation of ordinal categorical parameters into continuous ones}

The same LOO method was used to study the normalization and transformation of categorical parameters into continuous ones on the network imputation accuracy. Comparison of imputation accuracy (missing values imputation quality) can be seen in Table \ref{Period_table}.

\label{normalization_and_ordinal_results}
\begin{table}
\centering
\caption{Comparison of the mean NRMSE and accuracy of BNs, based on normalized Gross, Net Pay and Permeability parameters as well as converted Period into continuous . The best values are shown in bold.}
\label{Period_table}
\begin{tabular}{|l|l|l|}
\hline
\multicolumn{1}{|c|}{\textbf{Data preprocessing}} & \multicolumn{1}{c|}{\textbf{Mean NRMSE}} & \multicolumn{1}{c|}{\textbf{Mean Accuracy}} \\ \hline
No parameter changes     & 0.1775           &0.5175             \\ \hline
Period continuous        & 0.1787           &0.4275             \\ \hline
Log Period continuous    & \textbf{0.175}   & 0.47              \\ \hline
Log Gross                & 0.17625          &\textbf{0.5225}    \\ \hline
Log Net Pay               & 0.1962           &\textbf{0.5225}    \\ \hline
Log Permeability         & 0.18             &0.52               \\ \hline
\end{tabular}
\end{table}

The comparison results show that parameter transformation has little effect on the accuracy of the prediction. In converting the Period from categorical to continuous, the network prediction accuracy drops for both mean NRMSE and mean accuracy. By normalizing the continuous Period value through the logarithm, the accuracy of the network improves. Mean NRMSE improves over the baseline, but the mean accuracy falls short of the baseline. Net Pay and Permeability normalization improve mean accuracy but worsen mean NRMSE. Moreover, only the Gross normalization pays off. It improves well the mean accuracy and slightly loses on mean NRMSE.

\subsubsection{Bayesian networks built on analogues} \label{number_of_analogs}
In estimating geological parameters based on the nearest analogues reservoirs, the question arises as to how many nearby reservoirs would be sufficient for an accurate assessment. The following experiment was carried out for three distance metrics - cosine, Gower and Hamming:

\begin{enumerate}
 \item The target reservoir is selected from the database;
 \item The value of the geological parameter of interest to us is deleted;
 \item Then N of the nearest reservoirs is searched for based on the remaining parameter values. In this case, N takes the values 20, 40, 60 and 100;
 \item Based on the located nearby reservoirs, a network is built, and the remote value of the parameter is restored from the resulting network;
 \item Then, the imputation error is considered for networks built on different amounts of N.
\end{enumerate}

\begin{figure}
\begin{subfigure}[b]{.47\textwidth}
  \centering
  \includegraphics[width=1\linewidth]{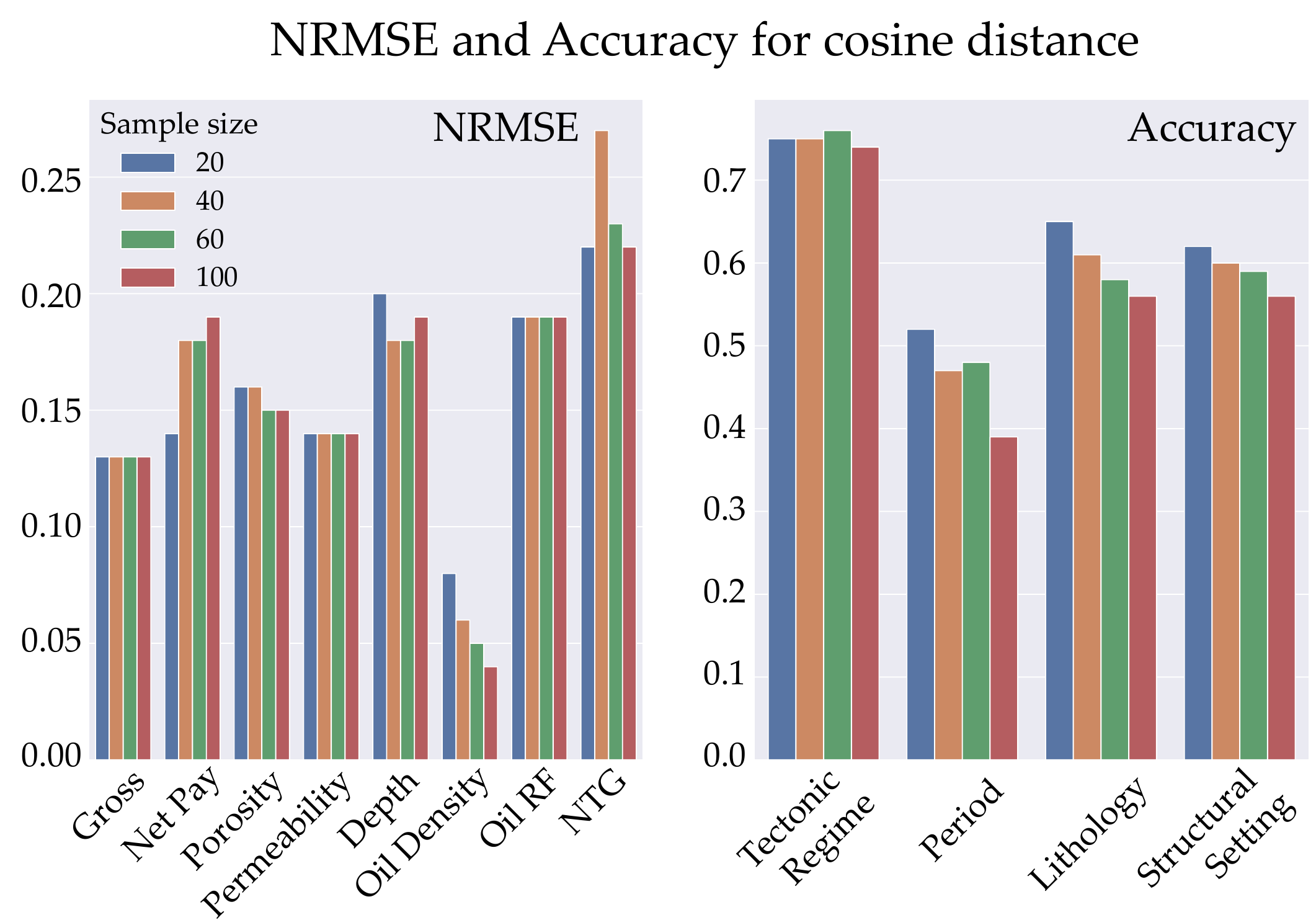}\\a)
\end{subfigure}%
\begin{subfigure}[b]{.47\textwidth}
  \centering
  \includegraphics[width=1\linewidth]{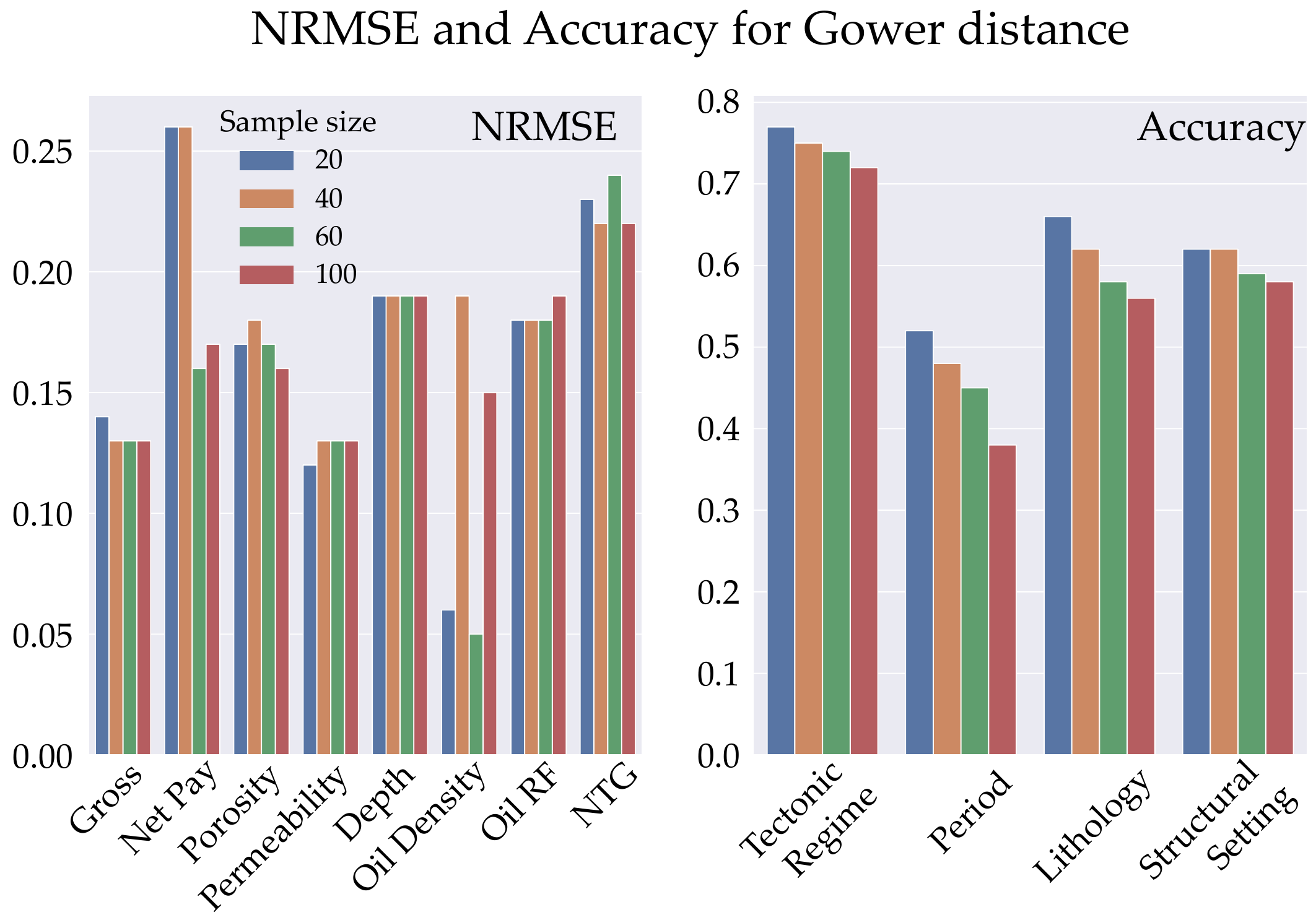}\\b)
\end{subfigure}
\begin{subfigure}[b]{.47\textwidth}
  \centering
  \includegraphics[width=1\linewidth]{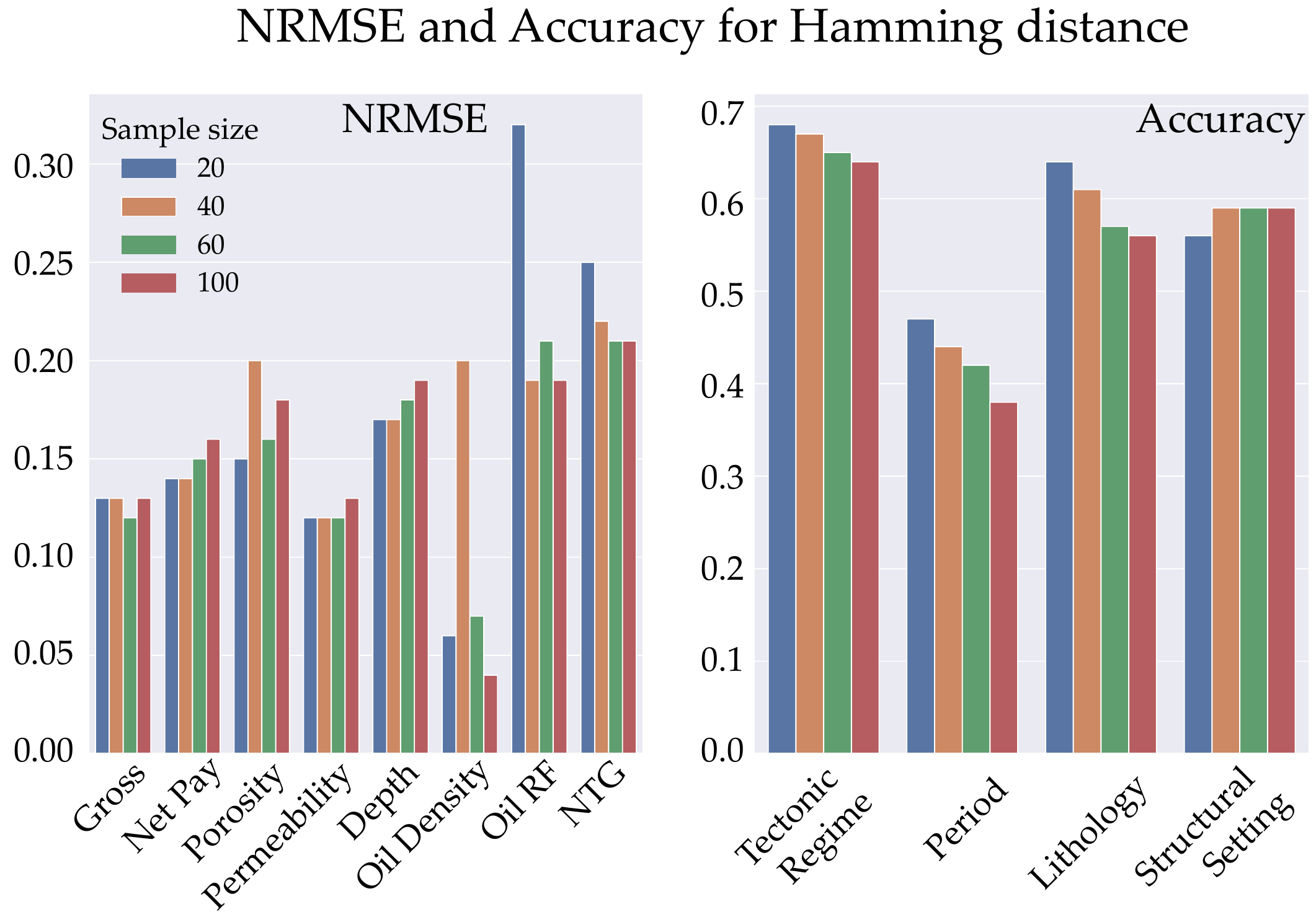}\\c)
\end{subfigure}
\caption{The results of reconstructing the parameters on networks with a different number of nearby analogues reservoirs, found on the basis of the cosine distance (a); the Gower distance (b); the Hamming distance (c). 
\label{size_res}}
\end{figure}

Fig. \ref{size_res} shows the results of the imputation error for all parameters and different sizes of the nearest reservoirs. We can see that there is an optimal number of nearby reservoirs for each parameter and distance metric. To determine, on average, for all parameters, the optimal amount, we calculate the average error and average accuracy for each number of the nearest reservoirs. Table \ref{sample_size} shows the average error and accuracy for a different number of nearby reservoirs, the optimal result for continuous and discrete parameters is highlighted in bold.

\begin{table}
\centering
\caption{Average error and accuracy for a different number of nearby reservoirs.
\label{sample_size}}
\begin{tabular}{|l|l|l|}
\hline
\multicolumn{3}{|c|}{\textbf{Cosine metric}}                                                                                 \\ \hline
\multicolumn{1}{|c|}{\textbf{Size}} & \multicolumn{1}{c|}{\textbf{Mean NRMSE}} & \multicolumn{1}{c|}{\textbf{Mean Accuracy}} \\ \hline
20                                  & 0.158                                    & \textbf{0.635}                              \\ \hline
40                                  & 0.164                                    & 0.608                                       \\ \hline
60                                  & \textbf{0.156}                           & 0.602                                       \\ \hline
100                                 & \textbf{0.156}                           & 0.562                                       \\ \hline
\multicolumn{3}{|c|}{\textbf{Gower metric}}                                                                                  \\ \hline
20                                  & 0.169                                    & \textbf{0.643}                              \\ \hline
40                                  & 0.185                                    & 0.618                                       \\ \hline
60                                  & \textbf{0.156}                           & 0.59                                        \\ \hline
100                                 & 0.168                                    & 0.56                                        \\ \hline

\multicolumn{3}{|c|}{\textbf{Hamming metric}}                                                                                  \\ \hline
20                                  & 0.168                                    & \textbf{0.588}                              \\ \hline
40                                  & 0.171                                    & 0.678                                       \\ \hline
60                                  & \textbf{0.152}                           & 0.558                                        \\ \hline
100                                 & 0.154                                    & 0.542                                        \\ \hline
\end{tabular}
\end{table}

Having chosen the optimal number of the closest analogues, equal to 60, we compare the results of restoring parameters on the entire network and clusters of analogues. For this, LOO validation was carried out, the results of which are presented in Table \ref{distance}.

\begin{table}
\centering
\caption{Average error and accuracy for parameters imputation with BN on all dataset and distance clusters (Highest accuracy and lowest error values are bold in each row).
\label{distance}}
\begin{tabular}{|c|c|c|c|c|}
\hline
\textbf{Parameter}  & \textbf{All data} & \textbf{Cosine metric} & \textbf{Gower metric} & \textbf{Hamming metric} \\ \hline
\multicolumn{5}{|c|}{\textbf{Accuracy}}                                                                            \\ \hline
Tectonic Regime     & \textbf{0.81}     & 0.76                   & 0.74                  & 0.65                    \\ \hline
Period              & 0.32              & \textbf{0.48}          & 0.45                  & 0.42                    \\ \hline
Lithology           & 0.56              & \textbf{0.58}          & \textbf{0.58}         & 0.57                    \\ \hline
Structural Setting  & 0.41              & \textbf{0.59}          & \textbf{0.59}         & \textbf{0.59}           \\ \hline
\multicolumn{5}{|c|}{\textbf{NRMSE}}                                                                               \\ \hline
Gross               & 0.14              & 0.13                   & 0.13                  & \textbf{0.12}           \\ \hline
Net Pay              & 0.18              & 0.18                   & 0.16                  & \textbf{0.15}           \\ \hline
Porosity            & 0.21              & \textbf{0.15}          & 0.17                  & 0.16                    \\ \hline
Permeability        & 0.14              & 0.14                   & 0.13                  & \textbf{0.12}           \\ \hline
Depth               & 0.19              & \textbf{0.18}          & 0.19                  & \textbf{0.18}           \\ \hline
Oil Density         & 0.15              & \textbf{0.05}          & \textbf{0.05}         & 0.07                    \\ \hline
Oil RF & 0.19              & 0.19                   & \textbf{0.18}         & 0.21                    \\ \hline
NTG                 & 0.22              & 0.22                   & 0.24                  & \textbf{0.21}           \\ \hline
\end{tabular}
\end{table}

The best result for discrete parameters can be observed for analogues found at cosine distance. Analogues give the best result for continuous parameters at the Hamming distance. In general, the use of analogues shows an improvement in the result of parameter prediction.

\subsubsection{Clustering reservoirs} \label{quality_clust}

\begin{figure}
\centering
    \includegraphics[width=1\linewidth]{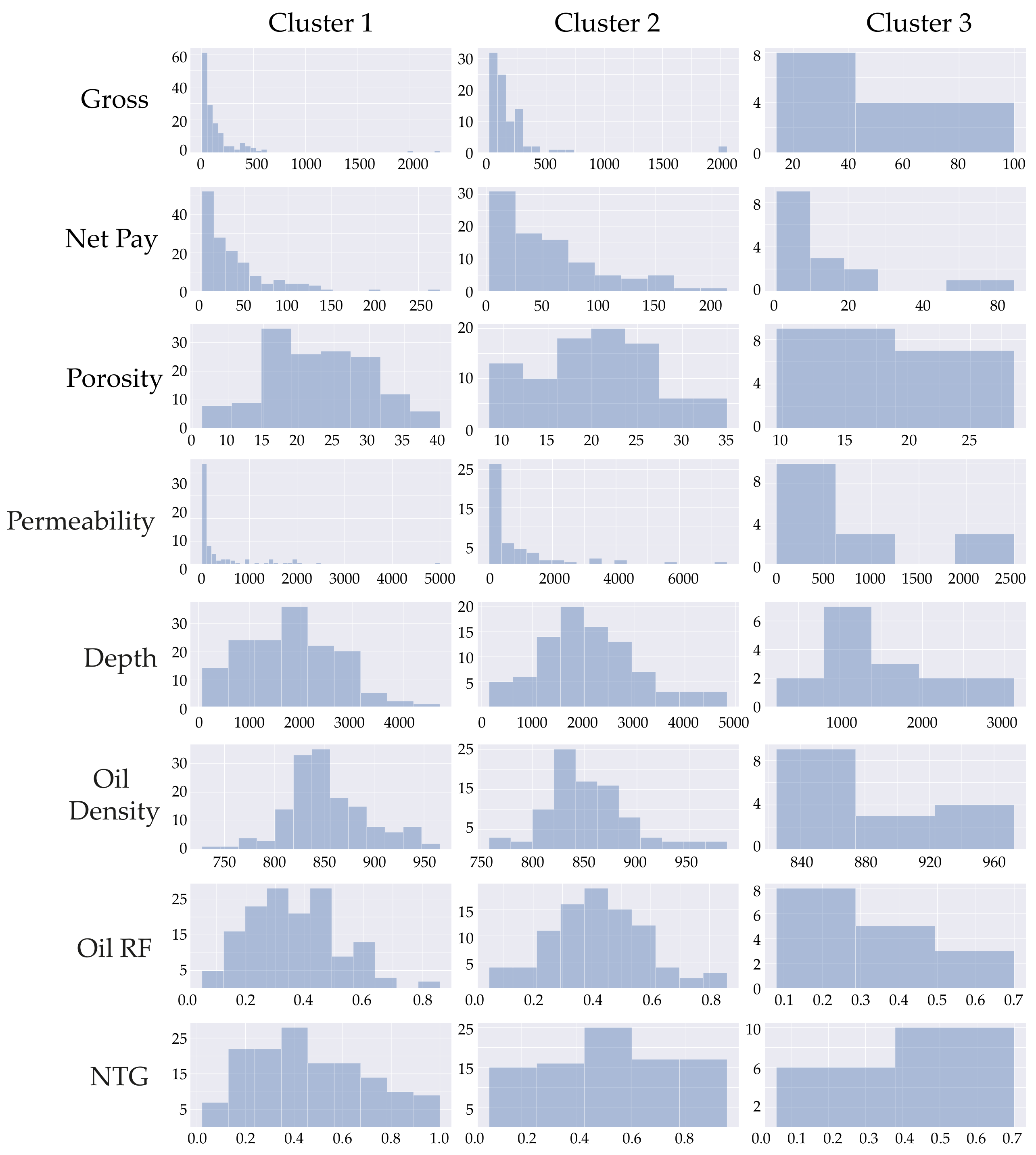}
    \caption{Distribution of continuous parameters within each cluster. \label{cluster_cont}}
\end{figure}

\begin{figure}
\centering
\includegraphics[width=1\linewidth]{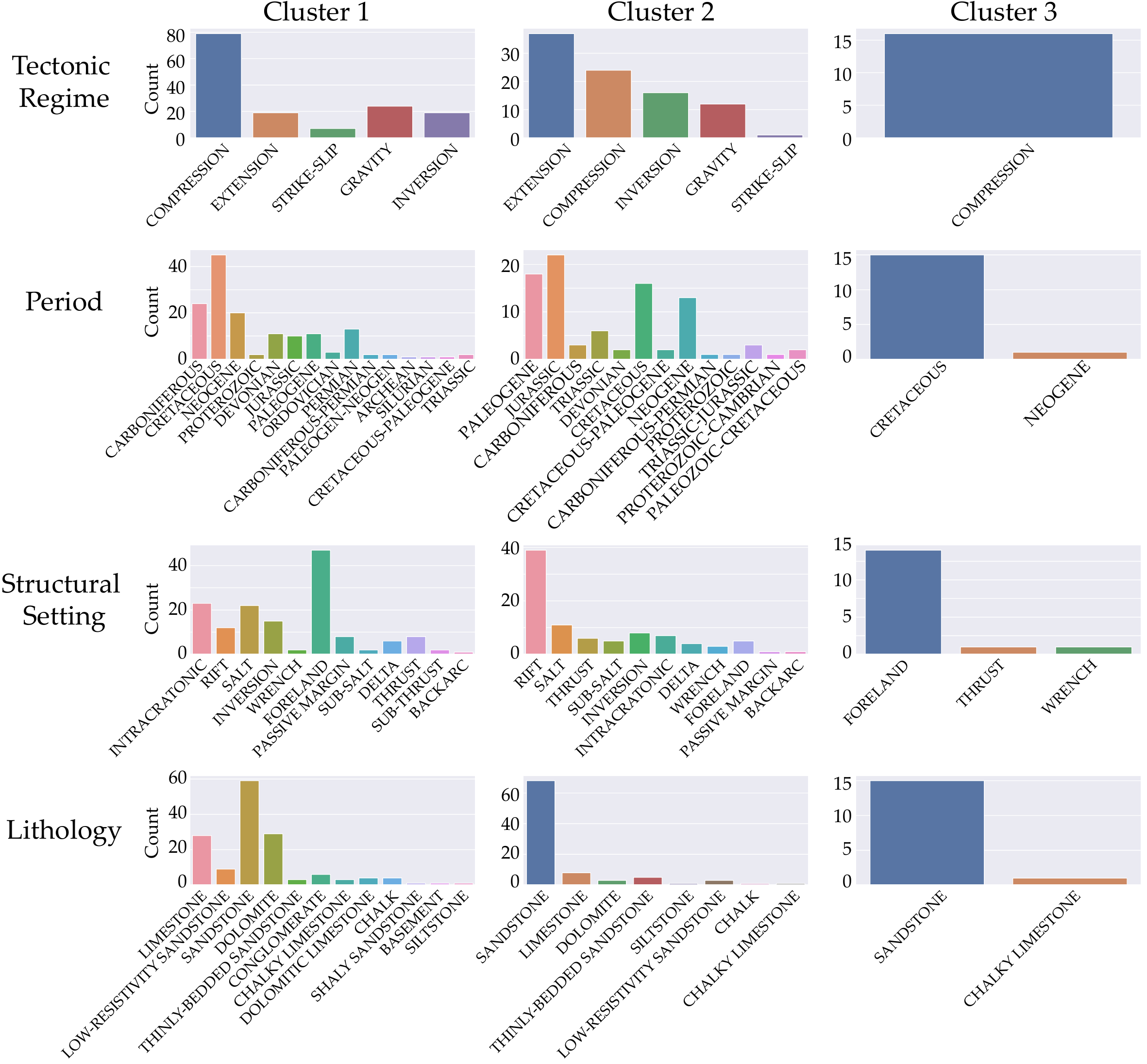}
\caption{Distribution of discrete parameters within each cluster. \label{cluster_disc}}
\end{figure}

To assess the quality of the clustering algorithm based on BNs (fig. \ref{bn_clustering}), the following experiment was carried out. The data were divided into training and test samples in a ratio of 80\% to 20\%. Then, based on the Gower distance, the 60 nearest reservoirs were selected on the training sample for each reservoir. The Gower distance was chosen because it best computes distances on mixed data. Networks were built on the obtained subsample, and their structures were clustered based on the Hamming distance. As a result, 3 clusters were allocated. fig. \ref{cluster_cont} and fig. \ref{cluster_disc} show the distributions of continuous and discrete parameters within each cluster.

The parameters were restored within the network built on the entire dataset and within the networks built on clusters on the test data set. Table \ref{table_clustering} shows a comparison of the accuracy and error of restoring parameters based on the entire database and based on clusters. It can be seen that prediction based on clusters of similar reservoirs has higher accuracy and lower error.

\begin{table}
\centering
\caption{Comparison of prediction results (missing values imputation) based on a network built on all data and based on networks built on reservoir clusters.
\label{table_clustering}}
\begin{tabular}{|c|c|c|}
\hline
\multicolumn{3}{|c|}{\textbf{Accuracy for discrete parameters}}       \\ \hline
\textbf{Parameter}  & \textbf{All dataset} & \textbf{With clustering} \\ \hline
Tectonic Regime     & \textbf{0.8}         & 0.78                     \\ \hline
Period              & 0.25                 & \textbf{0.33}            \\ \hline
Structural Setting  & \textbf{0.58}        & \textbf{0.58}            \\ \hline
Lithology           & 0.4                  & \textbf{0.49}            \\ \hline
\multicolumn{3}{|c|}{\textbf{NRMSE for continuous parameters}}        \\ \hline
Gross               & 0.18                 & \textbf{0.17}            \\ \hline
Net Pay              & 0.23                 & \textbf{0.21}            \\ \hline
Porosity            & \textbf{0.17}        & \textbf{0.17}            \\ \hline
Permeability        & 0.23                 & \textbf{0.19}            \\ \hline
Depth               & 0.25                 & \textbf{0.24}            \\ \hline
Oil Density         & 0.2                  & \textbf{0.18}            \\ \hline
Oil RF & 0.23                 & \textbf{0.19}            \\ \hline
NTG                 & 0.26                 & \textbf{0.22}            \\ \hline
\end{tabular}
\end{table}

\subsubsection{Filtering reservoirs}
To evaluate the quality of the filtering-based clustering algorithm, an experiment similar to the one presented earlier in fig. \ref{quality_clust} was conducted. Because of the large number of possible choices, the data were divided into training and test samples in the ratio of 90\% to 10\%. The data were filtered on the training sample for each variable, and then networks were built on them. Dendrograms were then constructed based on the Hamming distance between networks. Finally, clusters were constructed by cutting dendrograms across all possible thresholds.

For imputation, a cluster is first defined for the object to which it belongs according to the value of the filtering variable. The missing value is calculated in the standard way, using the BN built on this cluster. Thus, the peculiarity of filtering clusters is that they cannot predict the values of the variable used for filtering. Because of this, the value of accuracy and NRMSE for such variable was considered equal to the corresponding value for the network on full data obtained in the same way. 

Several possible clusters can be obtained for each filtering variable. We give preference to those that have the lowest mean NRMSE since we will need to predict a quantitative variable in fig. \ref{RF}. Tables \ref{filter_acc}, \ref{filter_nrmse} show the imputation performance on such clustering for each filtering variable. There are not only NRMSE values for all variables but also accuracy since the reservoir data in fig. \ref{RF} also contains a categorical value gap. For compactness of presentation, some attributes are reduced to abbreviations; their explanations are given in fig. \ref{abbr}. 

The main interest for us is those clusters that, on average, outperform the results shown by the network on the full dataset. The results for full networks are also given in the last rows. Note that the result differs from similar ones from fig. \ref{number_of_analogs} and fig. \ref{quality_clust}  because of differences in the test sample size. However, it is suitable for comparison with filter clusters because their quality was evaluated on the same test subsample.

\begin{table}
\centering
\caption{Accuracy for each variable and mean accuracy for different filtering clusters. 
\label{filter_acc}}
\begin{tabular}{|c|c|c|c|c|c|}
\hline
\multirow{2}{*}{\textbf{Filter}} &	\multicolumn{4}{|c|}{\textbf{Accuracy}}  &  \multirow{2}{*}{\textbf{Mean}}\\ \cline{2-5}
 &	TR	& Period &	Lithology &	SS &  \\ \hline
TR &	0.469 &	0.667 &	0.033 &	0.767 &	0.484 \\ \hline
Period &	\textbf{0.917} & 0.469 &	0.583 &	0.833 &	\textbf{0.700}  \\ \hline
Lithology &	0.647 &	0.765 &	0.469 &	0.235 &	0.529 \\ \hline
Gross &	0.613 &	0.613 &	0.452 &	0.484 &	0.540 \\ \hline
Net Pay &	0.690 &	0.621 &	0.483 &	0.310 &	0.526 \\ \hline
Por &	0.524 &	0.810 &	0.286 &	0.286 &	0.477 \\ \hline
Perm &	0.200 &	0.733 &	\textbf{0.600} &	0.133 &	0.417 \\ \hline
SS &	0.786 &	0.469 &	0.469 &	0.312 &	0.509 \\  \hline
Depth &	0.500 &	0.633 & 0.467 &	0.400 &	0.500 \\  \hline
OilD &	0.233 &	0.533 &	0.467 &	0.267 &	0.375 \\ \hline
RF &	0.357 &	\textbf{0.821} & 0.500 & 0.286 & 0.491 \\  \hline
NTG &	0.094 &	0.594 &	0.406 &	\textbf{0.938} &	0.508 \\  \hline \hline
Full &	0.469 &	0.469 &	0.469 &	0.312 &	0.430 \\  \hline
\end{tabular}
\end{table}

\begin{table}
\centering
\caption{NRMSE for each variable and mean NRMSE for different filtering clusters. 
\label{filter_nrmse}}
\begin{tabular}{|c|c|c|c|c|c|c|c|c|c|}
\hline
\multirow{2}{*}{\textbf{Filter}} &	\multicolumn{8}{|c|}{\textbf{NRMSE}}  &  \multirow{2}{*}{\textbf{Mean}}\\ \cline{2-9}
 & Gross &NP &	Por &	Perm &	Depth &	OilD &	RF &	NTG & \\ \hline
TR &	0.137 &	\textbf{0.079} & \textbf{0.080} &	0.180 &	0.143 &	0.023 &	0.145 &	0.185 &	\textbf{0.122} \\ \hline
Period &	0.285 &	0.280 &	0.371 &	0.377 &	0.179 &	0.039 &	0.316 &	0.520 &	0.296 \\ \hline
Lithology &	0.300 &	0.238 &	0.171 &	0.232 &	0.253 &	0.033 &	0.271 &	0.429 &	0.241 \\ \hline
Gross &	0.232 &	0.181 &	0.178 &	0.189 &	0.167 &	0.033 &	0.204 &	0.201 &	0.173 \\ \hline
NP &	0.193 &	0.187 &	0.142 &	0.617 &	0.286 &	\textbf{0.013} &	0.195 &	\textbf{0.128} &	0.220 \\ \hline
Por &	0.253 &	0.250 &	0.159 &	0.368 &	0.288 &	0.083 &	0.235 &	0.404 &	0.255\\ \hline
Perm &	0.202 &	0.152 &	0.460 &	0.172 &	0.211 &	0.045 &	0.360 &	0.337 &	0.242\\ \hline
SS &	\textbf{0.116} &	0.241 &	0.154 &	0.255 &	0.314 &	0.039 &	\textbf{0.132} &	0.204 &	0.182\\ \hline
Depth &	0.186 &	0.189 &	0.133 &	\textbf{0.162} &	0.217 &	\textbf{0.013} &	0.189 &	0.218 &	0.163\\ \hline
OilD &	0.124 &	0.138 &	0.127 &	0.170 &	\textbf{0.128} &	0.030 &	0.173 &	0.174 &	0.133\\ \hline
RF &	0.293 &	0.201 &	0.185 &	0.201 &	0.315 &	0.035 &	0.228 &	0.417 &	0.234\\ \hline
NTG &	0.223 &	0.175 &	0.152 &	0.283 &	0.197 &	0.022 &	\textbf{0.133} &	0.314 &	0.187\\ \hline \hline
Full &	0.232 &	0.187 &	0.159 &	0.172 &	0.217 &	0.030 &	0.228 &	0.314 &	0.192\\ \hline 
\end{tabular}
\end{table}

Consider the last column of Table \ref{filter_nrmse} with the mean NRMSE value. The best clustering by this parameter will be by the Tectonic Regime. It NRMSE evaluates all quantitative variables well enough and also performs reasonably well on categorical data, at least better than the network on the full dataset. It contains only one cluster, which contains all data except for those with the 'STRIKE-SLIP' value. In essence, it is not clustering but rejecting anomalies. A small change in the data set can lead to a much better result in mean NRMSE and reasonable accuracy. It is this clustering that we will use when estimating the RF in \ref{RF}.

\subsection{Comparison of the results of the physical model and Bayesian Networks} \label{RF}

To validate the Bayesian networks-based proposed approach, we propose to compare the oil RF prediction results of the Watt synthetic reservoir physical simulation with the Bayesian networks approaches. A detailed description of the physical simulation parameters performed with the ECLIPSE black oil model can be found in fig. \ref{fluid_flow_simulation}.
To carry out experiments with the oil RF prediction based on BNs, a synthetic Watt reservoir was considered with the following parameters: {Lithology:'SANDSTONE', Period: 'JURASSIC', Structural Setting: 'WRENCH', Gross: 160, Net Pay: 120, Permeability: 221, Porosity: 17, Depth: 1566, Oil Density: 815.5, NTG: 0.76}. In addition, continuous and categorical parameters which characterize the Watt reservoir were estimated from the simulation model. In accordance with the approaches discussed above, we will consider four approaches to predicting the oil RF parameter:
\begin{enumerate}
    \item Prediction based on a network built on the entire database;
    \item Prediction based on a network built on N nearest analogues reservoirs;
    \item Prediction based on a network built on filtered parameters
    \item Prediction based on a network built on a cluster of reservoirs.
\end{enumerate}

\begin{figure}
\centering
\begin{subfigure}[b]{.45\textwidth}
  \centering
  \includegraphics[width=0.95\linewidth]{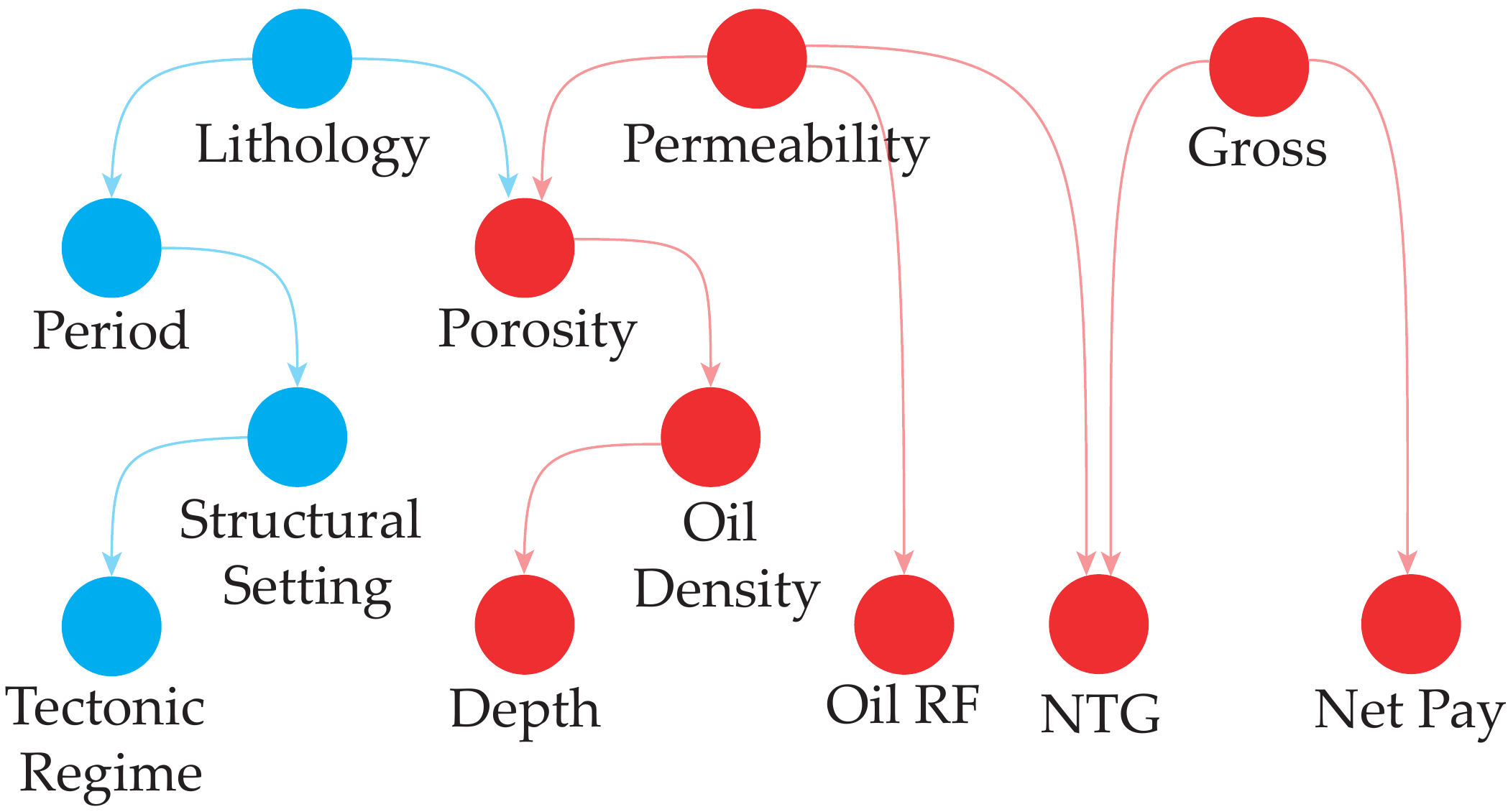}\\a)
\end{subfigure}%
\begin{subfigure}[b]{.45\textwidth}
  \centering
  \includegraphics[width=0.95\linewidth]{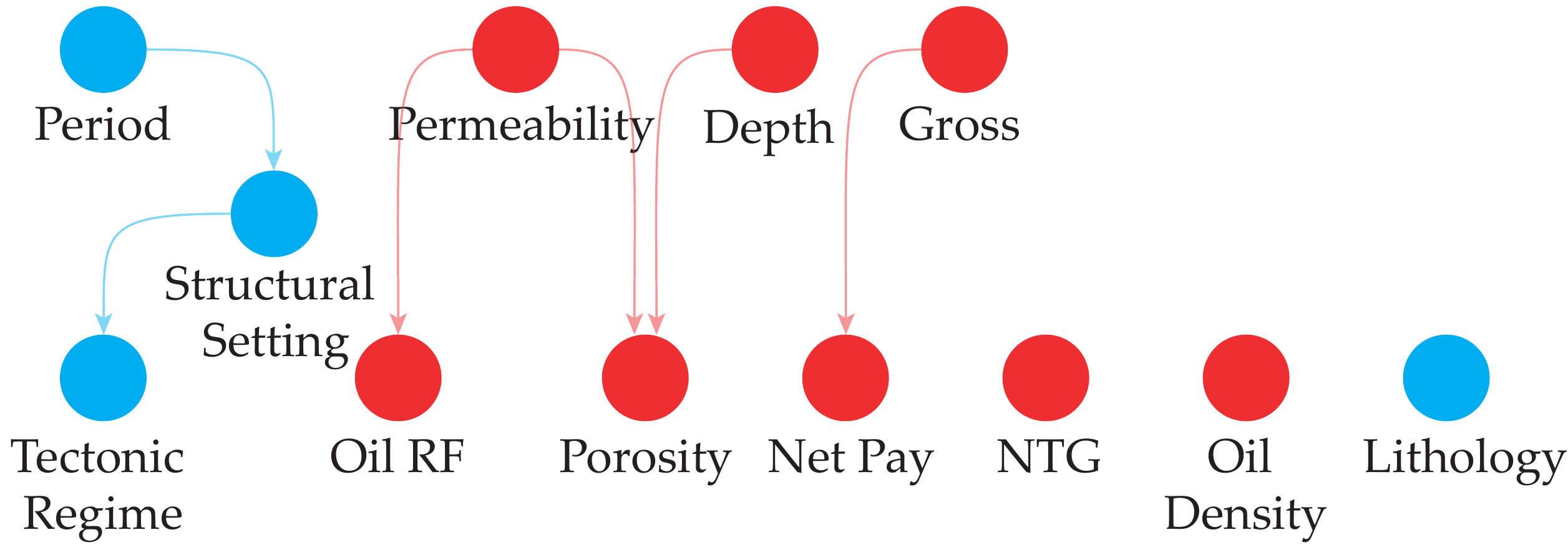}\\b)
\end{subfigure}
\begin{subfigure}[b]{.45\textwidth}
  \centering
  \includegraphics[width=0.95\linewidth]{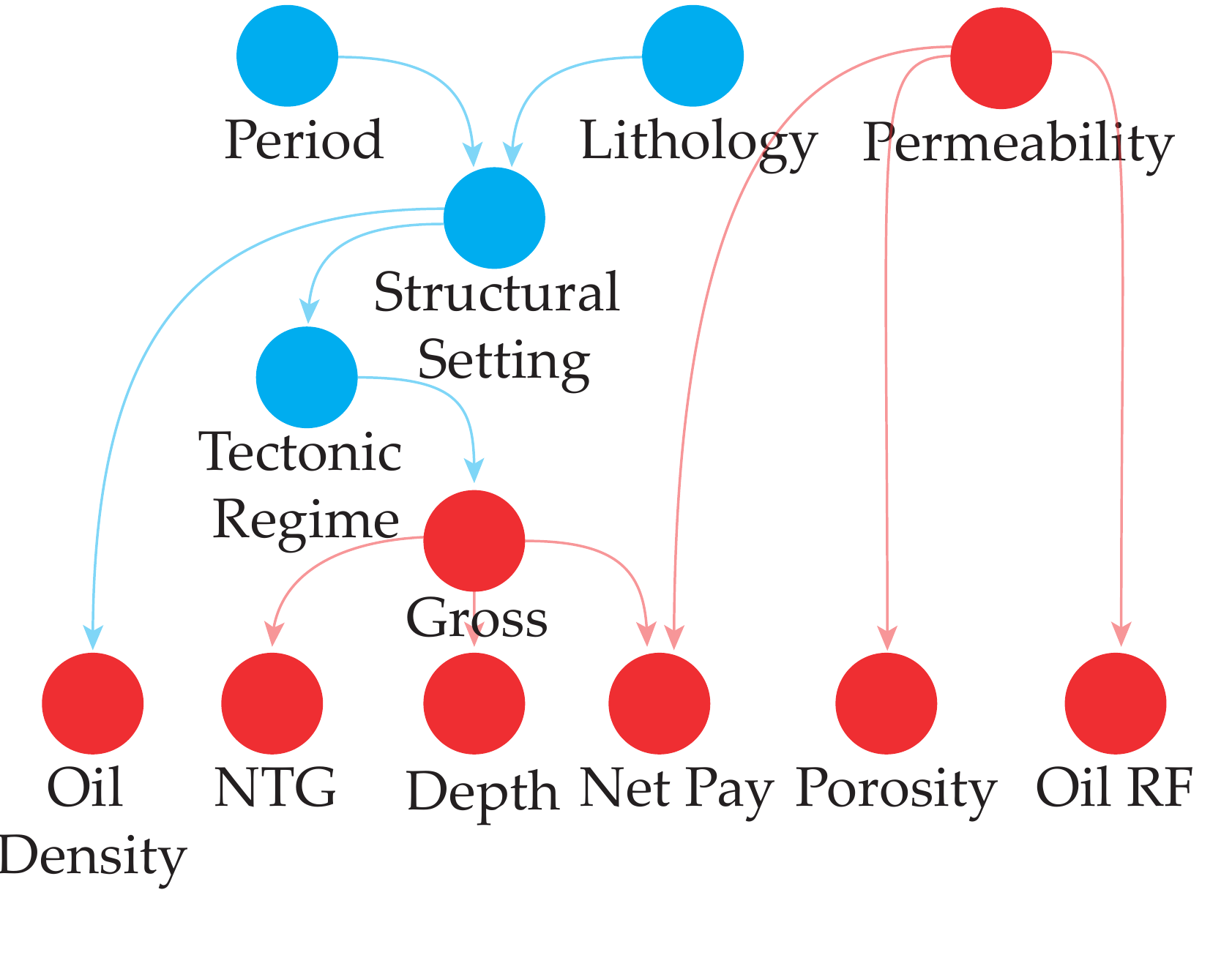}\\c)
\end{subfigure}
\begin{subfigure}[b]{.45\textwidth}
  \centering
  \includegraphics[width=0.95\linewidth]{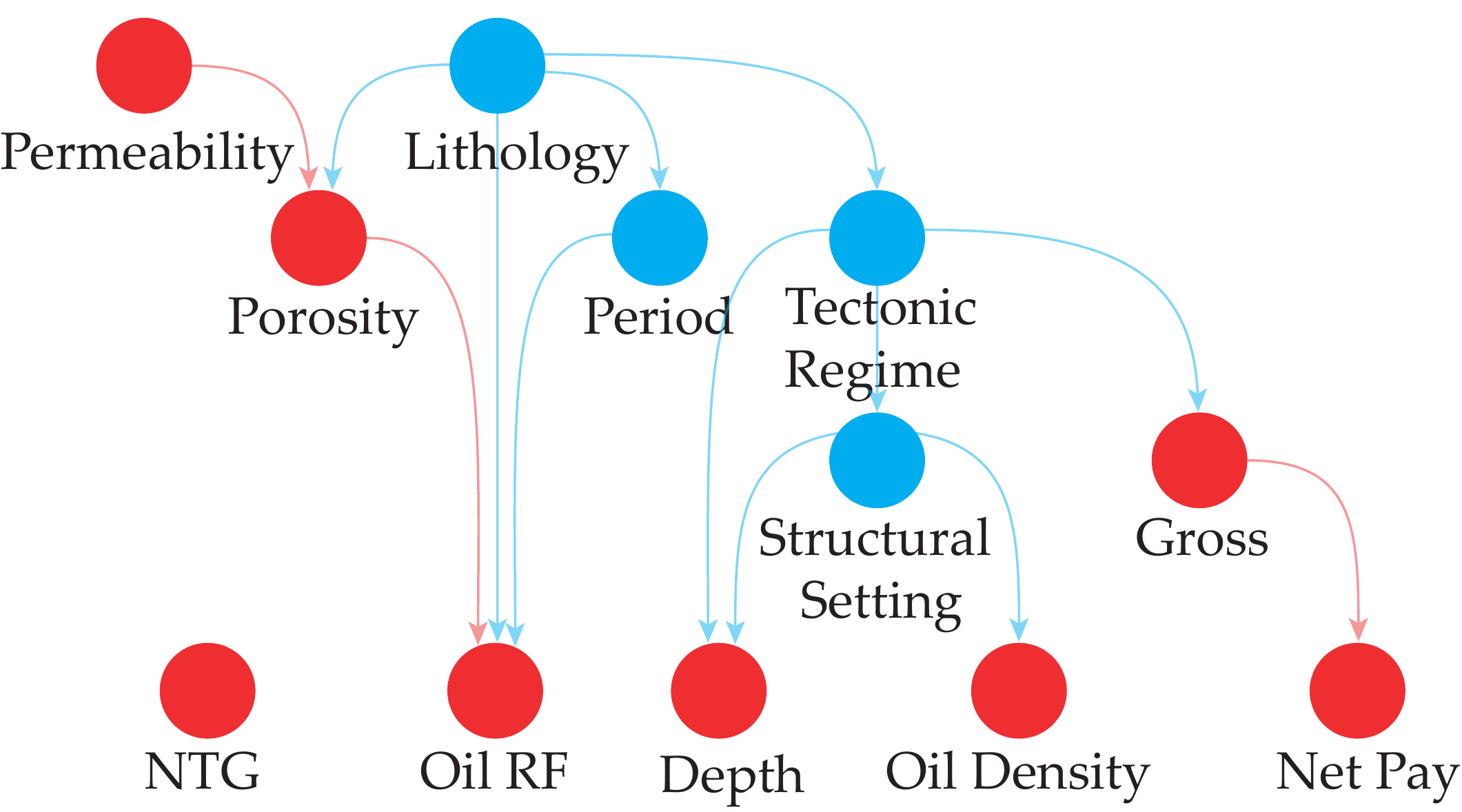}\\d)
\end{subfigure}
\begin{subfigure}[b]{.45\textwidth}
  \centering
  \includegraphics[width=0.95\linewidth]{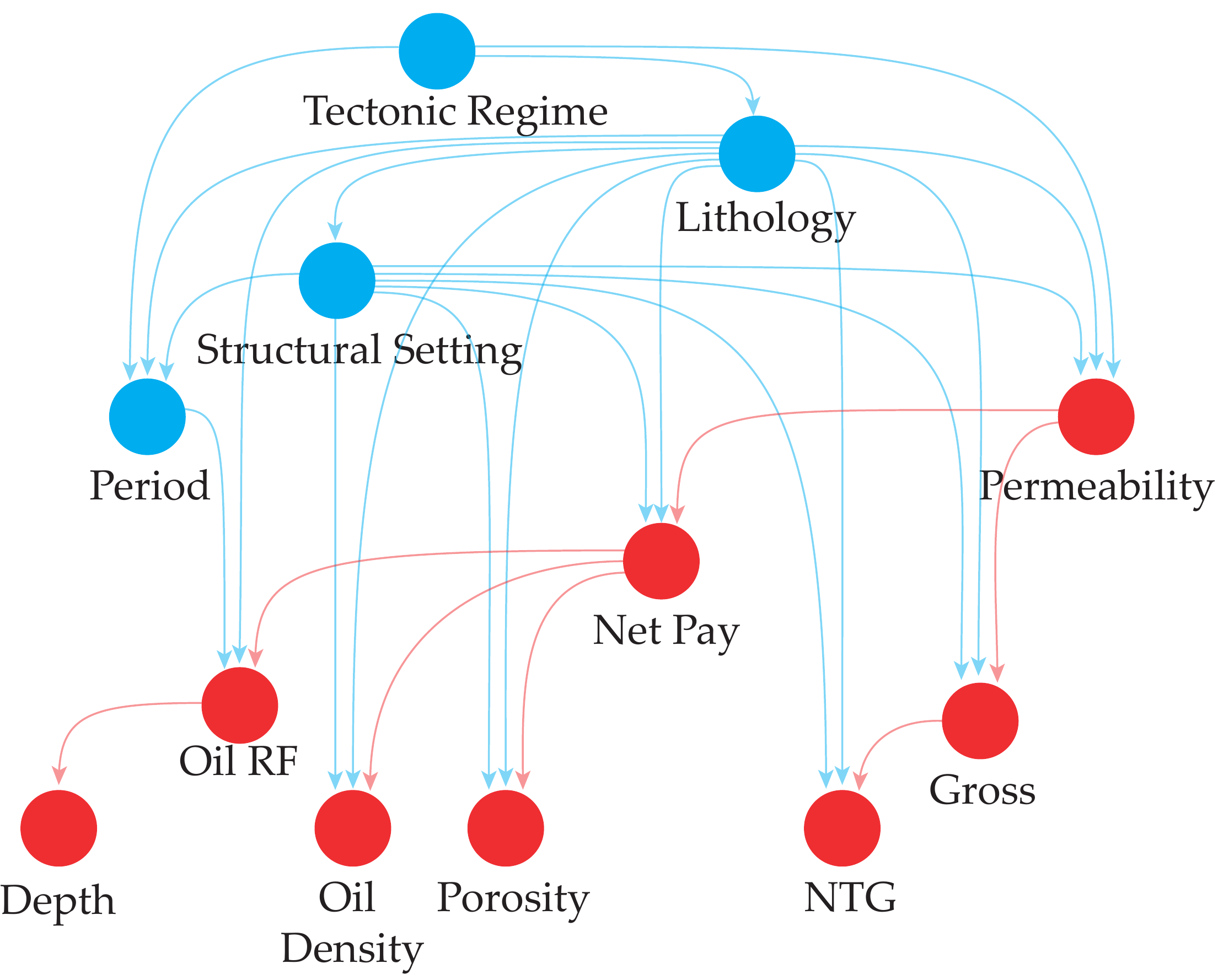}\\e)
\end{subfigure}
\begin{subfigure}[b]{.45\textwidth}
  \centering
  \includegraphics[width=0.95\linewidth]{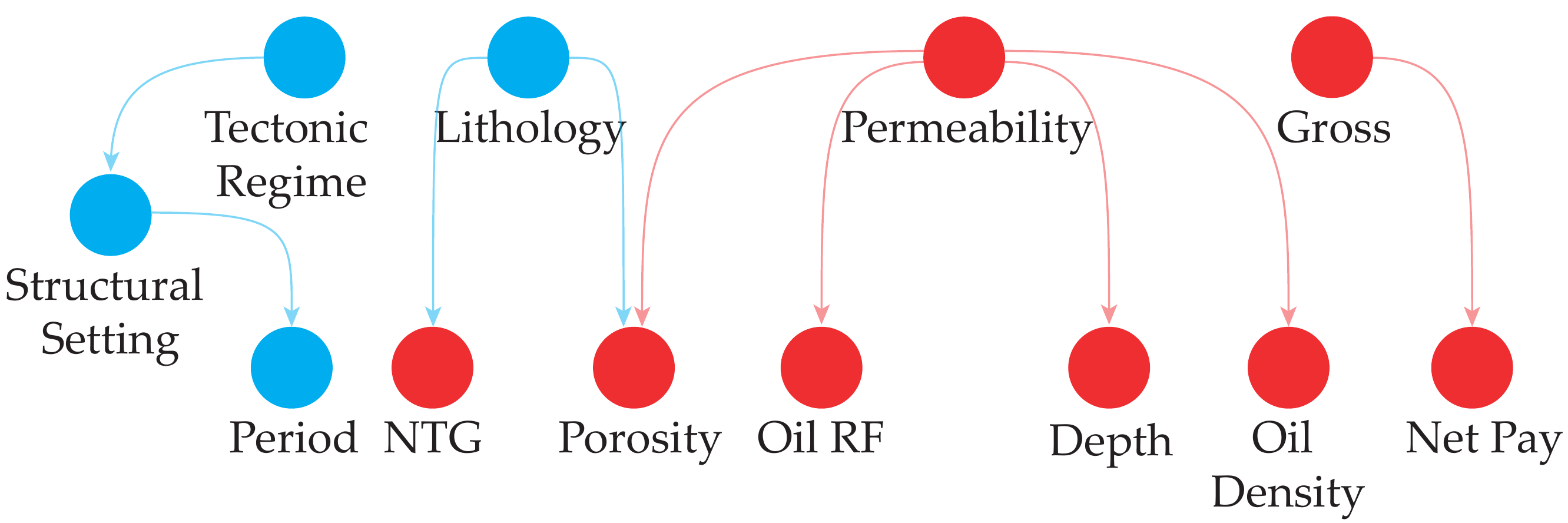}\\f)
\end{subfigure}
\caption{Different BN models: (a) full dataset BN; (b) cosine cluster BN; (c) Gower cluster BN; (d) Hamming cluster BN; (e) filtering parameters BN; (f) structure clustering BN. Blue - discrete parameters, red - continuous parameters.}
\label{fig:nets}
\end{figure}

A comparison of the predictions of the RF for the specified Watt reservoir on a physical model and various models of BNs is presented in Table \ref{table_RF}. Fig. \ref{fig:nets} shows different models of BNs structures obtained by different approaches. The relationships found by the algorithms in terms of available conditional distributions in data are shown in this figure. To be confident that revealed relations are relevant in geoscience interpretation, we have validated some of the robust edges in the networks by literature and fluid flow simulation model. First, there is a strong relationship (it is present almost at all possible variations of constructed BNs) between Lithology and Porosity or NTG. According to \cite{hawkes1998tectono} that is exactly the case within different depositional units of braided river basins. Second, the connection between Permeability and RF is evident on BNs. Also, this relation is relevant for the fluid flow simulation results we obtained (see Appendix \ref{fluid_flow_simulation}).

\begin{table}
\centering
\caption{The results of the mean value of the prediction of the RF for the physical model and for various models of BNs.
\label{table_RF}}
\begin{tabular}{|l|l|l|}
\hline
\multicolumn{1}{|c|}{\textbf{Model}} & \multicolumn{1}{c|}{\textbf{Mean RF}} & \multicolumn{1}{c|}{\textbf{95\% confidence interval}} \\ \hline
Physical model                       & 0.44                                  & 0.43 - 0.46                                            \\ \hline
BN on all dataset                    & 0.38                                  & 0.34 - 0.43                                            \\ \hline
BN on cosine cluster                 & 0.45                                  & 0.41 - 0.5                                             \\ \hline
BN on Gower cluster                  & 0.44                                  & 0.4 - 0.47                                             \\ \hline
BN on Hamming cluster                & 0.45                                  & 0.4 - 0.5                                              \\ \hline
BN on structure clustering           & 0.41                                  & 0.36 - 0.45                                            \\ \hline
BN on filtering cluster              & 0.46                                       & 0.43 - 0.5                                                        \\ \hline
\end{tabular}
\end{table}

\begin{figure}
\centering
    \includegraphics[width=1\linewidth]{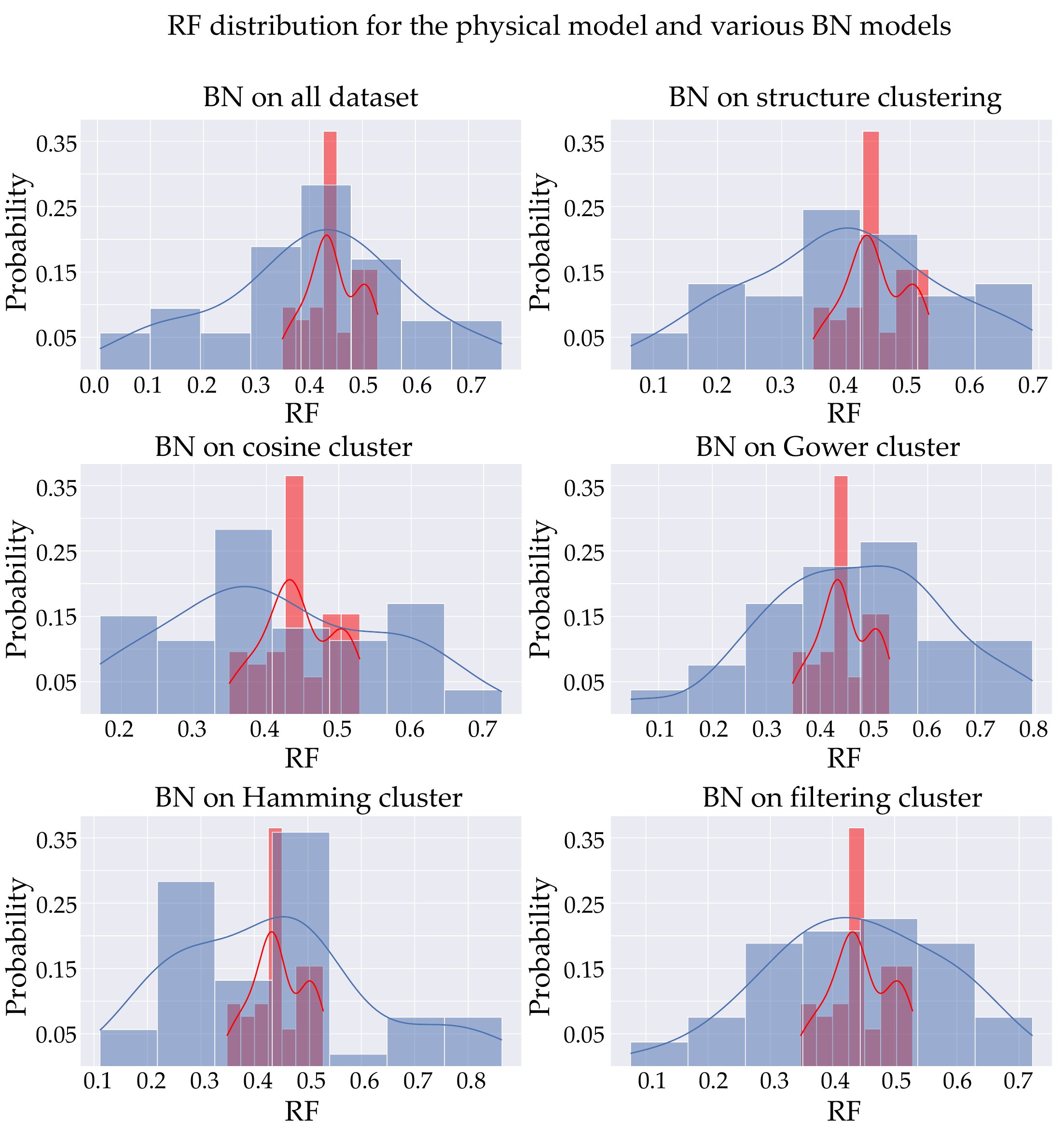}
    \caption{RF distributions for the physical model (red) and various BN models (blue). \label{rf_dist}}
\end{figure}

Fig. \ref{rf_dist} shows a comparison of the RF distributions of the physical model and models of BNs. It can be seen that the estimate using a BN has some uncertainty, this is due to the fact that the BN models the distribution seen in the data on which it was trained. Therefore, if the data contains an RF value, for example, 0.1 and 0.2, then they will be sampled in the network, albeit with small probabilities. The main thing is that the distribution modes of the BN and the simulator coincide, what can we say about the fact that the BN gives an adequate estimate of the parameter, but in a short time, on a small amount of data and with minimal involvement of expert knowledge. The physical simulation was performed in a commercial simulator in order to obtain the RF values for histograms. The following parameters were changed during stochastic modelling: top structure, fault models and their transmissibility, relative permeability model, variations in development strategy (default, with existing wells in the Watt dataset and tuned, with additional production wells). Also, different cut-off values obtained from facies logs in 6 exploration wells were used in facies modelling.  We performed petrophysical modelling for each facies model by ourselves using well log data. Other parameters such as grid size or the modelling approach were fixed. The total number of models is 108. It can be seen that the sampling from BNs has a large scatter, but the averages coincide with the model estimates, which indicates the adequacy of the estimate based on BN.

From Table \ref{table_RF}, one can conclude that BNs make it possible to predict the oil RF of the reservoir at the same level of confidence as the fluid flow model. According to \cite{bowman1993wytch} the recovery factor of the most similar type of braided river reservoir is also around 40\% which allow us to consider our results as a relevant approximation of the particular case. As can be seen from the table, the closest result was the result obtained based on the Gower distance. For this method, the coefficient of determination (R2) and the mean absolute error (MAE) were also calculated, equal to 0.3 and 0.12, respectively. Thus, a BN built on all data predicts a slightly lower oil RF. However, all the proposed approaches resulted closer to the physical model than the classical BN built on all data. Our results are in agreement with other work devoted to RF prediction with slightly higher values of R2 (the coefficient of determination) \cite{chicco2021coefficient} and MAE (mean absolute error) \cite{willmott2005advantages} which are equal to 0.38 and 0.09, respectively (LOO approach used for validating in both studies) \cite{makhotin2020machine}.

\section{Discussion}

Estimation of RF on early exploration stages may guide investment decision making. There have been several attempts to tackle this problem based solely on data-driven approaches. Mumtaz et al. \cite{mumtaz2020data} used different similarity measures, especially for categorical variables, to handle conceptual geological knowledge. Although the search for analogues reservoirs outperformed human selection, the authors claimed that there is no possibility of directly predicting the field's production and commercial outcomes based on reservoirs data only. A more optimistic view (Makhotin et al. \cite{makhotin2020machine}) includes using machine learning approaches (namely Gradient Boosting and Random Forest) to evaluate the RF-based on the database of more than 2000 reservoirs worldwide. Other possibilities to evaluate RF are to use production data in material balance, decline curve analysis or dimensionality reduction techniques. 
Time series analysis brings a new level of complexity in production forecasting and RF estimation. Busby \cite{busby2020deep} developed an enhanced approach based on decline curve analysis (DCA). DCA is more suitable for mature oilfields with a large number of producing wells. Physics informed neural networks are used to combine static and dynamic data for the robust forecast. Satija and Caers \cite{satija2017direct} also deal with time series dimensionality reduction to directly forecast production without fluid flow simulation. The combination of functional analysis and principal component analysis allowed authors to show relevant results compared to full-scale physical modelling.

The rapid development of hardware solutions and software methods in the field of machine learning leads to an unprecedented penetration of such solutions into all areas of science, business, and everyday life. Such techniques help in processing large amounts of multidimensional data with complex internal dependencies. Previously, a specialist with extensive experience and skills was required to operate with such a volume of information dependencies. Now machine learning methods can simplify this work by taking on statistical analysis, building models, processing such multivariate distributions, finding data dependencies, missing value prediction, target prediction, and much more. Our task is to use modern machine learning approaches to solve the urgent and high priority problems in the oil industry. One of these tasks is assessing the RF of the target reservoir since the importance of the RF is directly related to the economic feasibility of developing the reservoir. 

This paper considered the BNs approach to model the multivariate distribution of reservoir parameters. The approach supports the construction of an interpretable visual graphical model. To become a robust predictor, these models need to be guided by domain experts at the learning stage and result validation stage. Expert knowledge allows defining important relationships before structure and parameter learning and trimming unreasonable links after the result is obtained. However, it is possible to analyse multiple realisations of networks as the modelling process is measured in minutes, and the approach is extremely flexible. 

However, this approach has its drawbacks. For example, BNs do not work well with mixed data types, and reservoir parameters are precisely that. Since this approach is very sensitive to outliers in the data, high data quality is required. Additional processing of parameter values and uncertainties is needed. The approach also requires tuning hyperparameters for each new dataset.

This paper proposes a detailed pipeline to use BNs to model reservoir parameters. We show how to deal with shortcomings and also provide new approaches to increase the accuracy of models.

The first approach constructs BNs based on the parameters of analogues reservoirs. We considered three distance metrics used to estimate the proximity of reservoirs: cosine, Gower, and Hamming distances. We also compared four quantities of the closest analogues: 20, 40, 60 and 100. The 60 closest analogues turned out to be the optimal value. Regardless of the distance metric and numbers of analogues, models built on analogues give a more accurate result than those built on all data. 

The second approach clusters the reservoirs using BNs. For each reservoir, a BNs was built on the analogues of that reservoir. The resulting networks were then clustered. From our data, we established 3 clusters. Three BNs were built on these three clusters. The prediction accuracy on such networks was also higher than on the BN built on all data.

The third and final approach was the data filtering method. We defined a subsample of data by fixing a specific parameter with a certain value. Networks built on such subsamples can also show higher results than networks built on the entire dataset. This process mimics the logic of human interpreters who filter particular reservoir parameters to find suitable analogues across the globe or within the same basin. Therefore, the collection of such filtered networks represents a multilayer knowledge graph that would support professional decision-making \cite{hoffimann2021probabilistic}.

We would also like to note that one of the features of experiments carried out with BNs is that the results for each reservoir parameter may differ from all others. This leads to the possibility of choosing a specific learning algorithm and searching for analogues for each specific parameter.

All the proposed methods of using BNs for predicting parameters were tested experimentally using the example of a prediction of RF and showed their consistency between estimates obtained using BNs and the physical model. In general, BNs can be used to assess geological parameters, including the assessment of the RF. This approach is especially suitable for the early stages of reservoir development.

\vspace{6pt} 

\section{Conclusions}

We have developed an approach and python library suitable for RF estimation for a target reservoir based on a broad database of reservoirs in different petroleum basins. BNs are used for learning on mixed data (continuous and categorical parameters) to make a solid ground for decision support of domain experts. Graphical Probabilistic Models (represented in our paper by BNs) are actively used for explanatory analysis and causal inference and petroleum industry \cite{hoffimann2021probabilistic} and adjacent domains \cite{borsboom2021network}. We have validated our results for RF estimation on known relations from literature, by comparison with fluid flow model of braided river reservoir and we assume that our approach must be used in a constant feedback loop with professional geologists and reservoir engineers. Future plans include sensitivity analysis of BNs in order to evaluate the robustness of revealed relationships between reservoir parameters.\\

The following abbreviations are used in this manuscript:\\

\noindent 
\begin{tabular}{@{}ll}
\label{abbr}BN & Bayesian network\\
BIC & Bayesian information criterion\\
MI & Mutual information\\
NTG & Net Pay to Gross\\
TR & Tectonic Regime\\
Por & Porosity\\
Perm & Permeability\\
SS & Structural Setting\\
OilD & Oil Density\\
RF & Recovery factor\\
NP & Net Pay\\

\end{tabular}

\bibliographystyle{plain}
\bibliography{main}

\begin{thebibliography}{10}

\bibitem{ahmed2010reservoir}
T.~Ahmed.
\newblock Reservoir engineering handbook.
\newblock {\em Reservoir Engineering Handbook}, 01 2010.

\bibitem{al2016developed}
Omar Al-Fatlawi, Md~Hossain, Steven Hicks, and Ali Saeedi.
\newblock Developed material balance approach for estimating gas initially in
  place and ultimate recovery for tight gas reservoirs.
\newblock 11 2016.

\bibitem{andriushchenko2020analysis}
PD~Andriushchenko, IU~Deeva, AV~Kalyuzhnaya, AV~Bubnova, AG~Voskresenskiy, and
  NV~Bukhanov.
\newblock Analysis of parameters of oil and gas fields using bayesian networks.
\newblock In {\em Data Science in Oil \& Gas}, volume 2020, pages 1--10.
  European Association of Geoscientists \& Engineers, 2020.

\bibitem{ani2016reservoir}
Maureen Ani, Gbenga Oluyemi, Andrei Petrovski, Sina Rezaei-Gomari, et~al.
\newblock Reservoir uncertainty analysis: The trends from probability to
  algorithms and machine learning.
\newblock In {\em SPE Intelligent Energy International Conference and
  Exhibition}. Society of Petroleum Engineers, 2016.

\bibitem{arnold2013hierarchical}
Dan Arnold, Vasily Demyanov, Dominic Tatum, Mike Christie, T~Rojas, Sebastian
  Geiger, and P~Corbett.
\newblock Hierarchical benchmark case study for history matching, uncertainty
  quantification and reservoir characterisation.
\newblock {\em Computers \& Geosciences}, 50:4--15, 2013.

\bibitem{borsboom2021network}
Denny Borsboom, Marie~K Deserno, Mijke Rhemtulla, Sacha Epskamp, Eiko~I Fried,
  Richard~J McNally, Donald~J Robinaugh, Marco Perugini, Jonas Dalege, Giulio
  Costantini, et~al.
\newblock Network analysis of multivariate data in psychological science.
\newblock {\em Nature Reviews Methods Primers}, 1(1):1--18, 2021.

\bibitem{bottcher2001learning}
Susanne Bottcher.
\newblock Learning bayesian networks with mixed variables.
\newblock In {\em International Workshop on Artificial Intelligence and
  Statistics}, pages 13--20. PMLR, 2001.

\bibitem{bowman1993wytch}
MBJ Bowman, NM~McClure, and DW~Wilkinson.
\newblock Wytch farm oilfield: deterministic reservoir description of the
  triassic sherwood sandstone.
\newblock In {\em Geological Society, London, Petroleum Geology Conference
  series}, volume~4, pages 1513--1517. Geological Society of London, 1993.

\bibitem{bubnova2021}
A.~V. Bubnova, I.~Deeva, and A.~V. Kalyuzhnaya.
\newblock Mixbn: library for learning bayesian networks from mixed data.
\newblock {\em arXiv preprint arXiv:2106.13194}, 2021.

\bibitem{busby2020deep}
D~Busby.
\newblock Deep-dca a new approach for well hydrocarbon production forecasting.
\newblock In {\em ECMOR XVII}, volume 2020, pages 1--10. European Association
  of Geoscientists \& Engineers, 2020.

\bibitem{carvalho2009scoring}
Alexandra~M Carvalho.
\newblock Scoring functions for learning bayesian networks.
\newblock {\em Inesc-id Tec. Rep}, 12:1--48, 2009.

\bibitem{chicco2021coefficient}
Davide Chicco, Matthijs~J Warrens, and Giuseppe Jurman.
\newblock The coefficient of determination r-squared is more informative than
  smape, mae, mape, mse and rmse in regression analysis evaluation.
\newblock {\em PeerJ Computer Science}, 7:e623, 2021.

\bibitem{chickering2002optimal}
David~Maxwell Chickering.
\newblock Optimal structure identification with greedy search.
\newblock {\em Journal of machine learning research}, 3(Nov):507--554, 2002.

\bibitem{cooper1992bayesian}
Gregory~F Cooper and Edward Herskovits.
\newblock A bayesian method for the induction of probabilistic networks from
  data.
\newblock {\em Machine learning}, 9(4):309--347, 1992.

\bibitem{craft1992applied}
B.~Craft, M.~Hawkins, and Ronald Terry.
\newblock Applied petroleum reservoir engineering.
\newblock 01 1991.

\bibitem{dake1978fundamentsls}
L.P. Dake.
\newblock {\em Fundamentals of reservoir engineering}.
\newblock Elsevier Scientific Pub. Co. ; distributors for the U.S. and Canada
  Elsevier North-Holland, 1978.

\bibitem{BAMT}
Irina Deeva, Anna Bubnova, Petr Andriushchenko, and Nikolay~O. Nikitin.
\newblock Publicly available analogue of data, 2021.

\bibitem{deeva2021oil}
Irina Deeva, Anna Bubnova, Petr Andriushchenko, Anton Voskresenskiy, Nikita
  Bukhanov, Nikolay~O Nikitin, and Anna~V Kalyuzhnaya.
\newblock Oil and gas reservoirs parameters analysis using mixed learning of
  bayesian networks.
\newblock In {\em International Conference on Computational Science}, pages
  394--407. Springer, 2021.

\bibitem{gamez2011learning}
Jos{\'e}~A G{\'a}mez, Juan~L Mateo, and Jos{\'e}~M Puerta.
\newblock Learning bayesian networks by hill climbing: efficient methods based
  on progressive restriction of the neighborhood.
\newblock {\em Data Mining and Knowledge Discovery}, 22(1):106--148, 2011.

\bibitem{gomes2018benchmarking}
Jorge Gomes, Ram Narayanan, Humberto Parra, Luigi Saputelli, and Yogesh Bansal.
\newblock Benchmarking recovery factors for carbonate reservoirs: Key
  challenges and main findings from middle eastern fields.
\newblock 11 2018.

\bibitem{hawkes1998tectono}
PW~Hawkes, AJ~Fraser, and CCG Einchcomb.
\newblock The tectono-stratigraphic development and exploration history of the
  weald and wessex basins, southern england, uk.
\newblock {\em Geological Society, London, Special Publications},
  133(1):39--65, 1998.

\bibitem{hoffimann2021probabilistic}
J{\'u}lio Hoffimann, Sandro~Rama Fiorini, Breno de~Carvalho, Andres Codas,
  Carlos Raoni, Bianca Zadrozny, Rog{\'e}rio de~Paula, Oksana Popova, Maxim
  Mityaev, Irina Shishmanidi, et~al.
\newblock Probabilistic knowledge-based characterization of conceptual
  geological models.
\newblock {\em Applied Computing and Geosciences}, 10:100055, 2021.

\bibitem{jahn2003hydrocarbon}
Frank Jahn, M.~Cook, and M.~Graham.
\newblock {\em Hydrocarbon Exploration and Production}.
\newblock 01 2003.

\bibitem{jia2016novel}
L~Jia, A~John, N~Kumar, R~Bialas, TP~Lanson, and XD~Jing.
\newblock Novel benchmark and analogue method to evaluate heavy oil projects.
\newblock In {\em SPE Heavy Oil Conference and Exhibition}. OnePetro, 2016.

\bibitem{jolley2010reservoir}
S.~Jolley, Quentin Fisher, and R.B. Ainsworth.
\newblock Reservoir compartmentalization: An introduction.
\newblock {\em Geological Society of London Special Publications}, 347:1--8, 11
  2010.

\bibitem{koller2009probabilistic}
Daphne Koller and Nir Friedman.
\newblock {\em Probabilistic graphical models: principles and techniques}.
\newblock MIT press, 2009.

\bibitem{larue2005controversy}
D.~Larue and Francois Friedmann.
\newblock The controversy concerning stratigraphic architecture of channelized
  reservoirs and recovery by waterflooding.
\newblock {\em Petroleum Geoscience - PETROL GEOSCI}, 11:131--146, 05 2005.

\bibitem{mahmoud2019estimation}
Ahmed~Abdulhamid Mahmoud, Salaheldin Elkatatny, Weiqing Chen, and Abdulazeez
  Abdulraheem.
\newblock Estimation of oil recovery factor for water drive sandy reservoirs
  through applications of artificial intelligence.
\newblock {\em Energies}, 12:3671, 09 2019.

\bibitem{makhotin2020machine}
Ivan Makhotin, Denis Orlov, Dmitry Koroteev, Evgeny Burnaev, Aram Karapetyan,
  and Dmitry Antonenko.
\newblock Machine learning for recovery factor estimation of an oil reservoir:
  a tool for de-risking at a hydrocarbon asset evaluation.
\newblock {\em arXiv preprint arXiv:2010.03408}, 2020.

\bibitem{martin2013new}
Hilario Martin~Rodriguez, Elena Escobar, Sonia Embid, N~Rodriguez, Mohamed
  Hegazy, Larry~W Lake, et~al.
\newblock New approach to identify analogue reservoirs.
\newblock In {\em SPE Annual Technical Conference and Exhibition}. Society of
  Petroleum Engineers, 2013.

\bibitem{martinelli2013building}
Gabriele Martinelli, Jo~Eidsvik, Richard Sinding-Larsen, Sara Rekstad, and
  Tapan Mukerji.
\newblock Building bayesian networks from basin-modelling scenarios for
  improved geological decision making.
\newblock {\em Petroleum Geoscience}, 19(3):289--304, 2013.

\bibitem{masoudi2015feature}
Pedram Masoudi, Yousef Asgarinezhad, and Behzad Tokhmechi.
\newblock Feature selection for reservoir characterisation by bayesian network.
\newblock {\em Arabian Journal of Geosciences}, 8(5):3031--3043, 2015.

\bibitem{mazumder2021failure}
Ram~K Mazumder, Abdullahi~M Salman, and Yue Li.
\newblock Failure risk analysis of pipelines using data-driven machine learning
  algorithms.
\newblock {\em Structural Safety}, 89:102047, 2021.

\bibitem{muggeridge2013recovery}
Cockin Muggeridge, Frampton Webb, Moulds Collins, and Salino.
\newblock Recovery rates, enhanced oil recovery and technological limits.
\newblock 2013.

\bibitem{mumtaz2020data}
Summaya Mumtaz, Irina Pene, Adnan Latif, and Martin Giese.
\newblock Data-based support for petroleum prospect evaluation.
\newblock {\em Earth Science Informatics}, 13(4):1305--1324, 2020.

\bibitem{noureldien2015using}
A.H Noureldien, D.M.; El-Banbi.
\newblock Using artificial intelligence in estimating oil recovery factor.
\newblock {\em In Proceedings of the SPE North Africa Technology Conference and
  Exhibition}.

\bibitem{oil2017recovery}
Oil and Gas Authority.
\newblock Recovery factor benchmarking. uk continental shelf (ukcs) oilfields,
  9 2017.

\bibitem{popova2018analogy}
Oksana Popova et~al.
\newblock Analogy in the world of geological uncertainties, or how reservoir
  analogs may refine your probabilistic geomodel.
\newblock In {\em SPE Annual Caspian Technical Conference and Exhibition}.
  Society of Petroleum Engineers, 2018.

\bibitem{price1980crude}
Leigh~C. Price.
\newblock Crude oil degradation as an explanation of the depth rule.
\newblock {\em Chemical Geology}, 28:1--30, 1980.

\bibitem{rahuma2013prediction}
Khulud Rahuma, H.~Mohamed, N.~Hissein, and S.~Giuma.
\newblock Prediction of reservoir performance applying decline curve analysis.
\newblock {\em International Journal of Chemical Engineering and Applications},
  4:74--77, 01 2013.

\bibitem{satija2017direct}
Addy Satija, Celine Scheidt, Lewis Li, and Jef Caers.
\newblock Direct forecasting of reservoir performance using production data
  without history matching.
\newblock {\em Computational Geosciences}, 21(2):315--333, 2017.

\bibitem{scanagatta2019survey}
Mauro Scanagatta, Antonio Salmer{\'o}n, and Fabio Stella.
\newblock A survey on bayesian network structure learning from data.
\newblock {\em Progress in Artificial Intelligence}, 8(4):425--439, 2019.

\bibitem{shepherd2009factors}
Mike Shepherd.
\newblock {Factors Influencing Recovery from Oil and Gas Fields}.
\newblock 01 2009.

\bibitem{silva2018sensitivity}
R~Silva, L~Gualda, L~Lima, E~VitalBrazil, R~Cerqueiro, R~Paula, and U~Mello.
\newblock Sensitivity analysis in a machine learning methodology for reservoir
  analogues.
\newblock In {\em Rio Oil \& Gas Expo and Conference Proceedings}, 2018.

\bibitem{sircar2021application}
Anirbid Sircar, Kriti Yadav, Kamakshi Rayavarapu, Namrata Bist, and Hemangi
  Oza.
\newblock Application of machine learning and artificial intelligence in oil
  and gas industry.
\newblock {\em Petroleum Research}, 2021.

\bibitem{sloan2003quantification}
Rod Sloan.
\newblock {Quantification of Uncertainty in Recovery Efficiency Predictions:
  Lessons Learned from 250 Mature Carbonate Fields}.
\newblock All Days, 10 2003.
\newblock SPE-84459-MS.

\bibitem{sloan2004global}
Roderick Sloan and S.Q. Sun.
\newblock Global survey on use of geological analogs.
\newblock 31, 06 2004.

\bibitem{thanh2021integrated}
Hung~Vo Thanh and Yuichi Sugai.
\newblock Integrated modelling framework for enhancement history matching in
  fluvial channel sandstone reservoirs.
\newblock {\em Upstream Oil and Gas Technology}, 6:100027, 2021.

\bibitem{thanh2020application}
Hung~Vo Thanh, Yuichi Sugai, and Kyuro Sasaki.
\newblock Application of artificial neural network for predicting the
  performance of co 2 enhanced oil recovery and storage in residual oil zones.
\newblock {\em Scientific reports}, 10(1):1--16, 2020.

\bibitem{tyler1991architectural}
Noel Tyler and Robert~J. Finley.
\newblock {Architectural Controls on the Recovery of Hydrocarbons From
  Sandstone Reservoers}.
\newblock 01 1991.

\bibitem{voskresenskiy2020variations}
A~Voskresenskiy, M~Butorina, O~Popova, N~Bukhanov, Z~Filippova, R~Brandao,
  V~Segura, and E~Vital Brazil.
\newblock Variations in ranked list of reservoir analogs as an effect of search
  preferences.
\newblock In {\em Saint Petersburg 2020}, volume 2020, pages 1--5. European
  Association of Geoscientists \& Engineers, 2020.

\bibitem{wickens2010rapid}
Laurence~M Wickens and Roy Kelly.
\newblock Rapid assessment of potential recovery factor: a new correlation
  demonstrated on uk and usa fields.
\newblock In {\em SPE Annual Technical Conference and Exhibition}. OnePetro,
  2010.

\bibitem{willmott2005advantages}
Cort~J Willmott and Kenji Matsuura.
\newblock Advantages of the mean absolute error (mae) over the root mean square
  error (rmse) in assessing average model performance.
\newblock {\em Climate research}, 30(1):79--82, 2005.

\end{thebibliography}

\appendix
\section{Fluid flow simulation benchmark}
\label{fluid_flow_simulation}
\begin{figure}
\centering
    \includegraphics[width=1\linewidth]{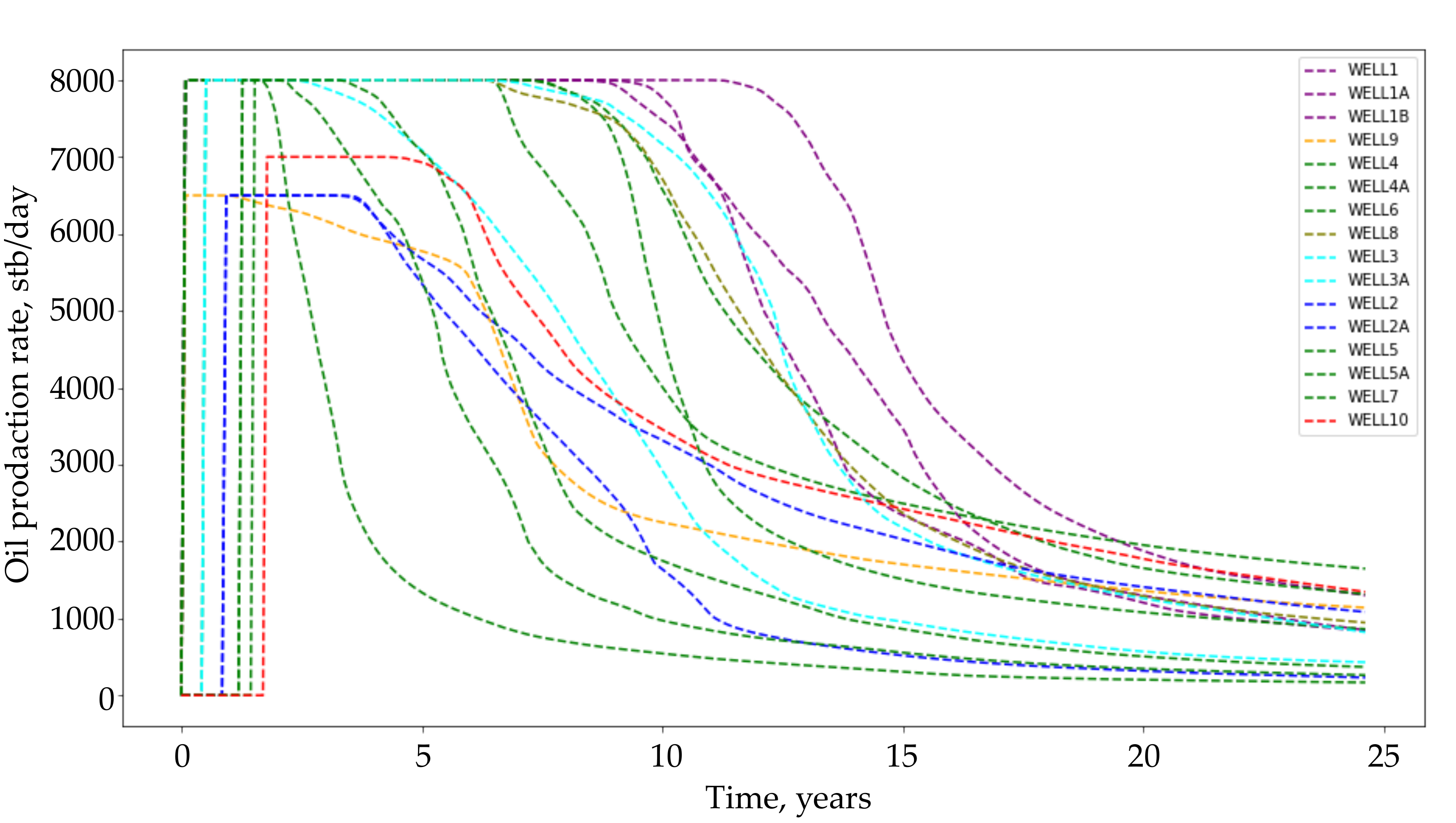}
    \caption{Oil production rates for each well simulated for the Watt benchmark reservoir (colours assigned according to zone division from Arnold et al \cite{arnold2013hierarchical}).\label{fig:WOPR}}
\end{figure}

\begin{figure}
\centering
\begin{subfigure}[b]{.55\textwidth}
  \centering
  \includegraphics[width=0.95\linewidth]{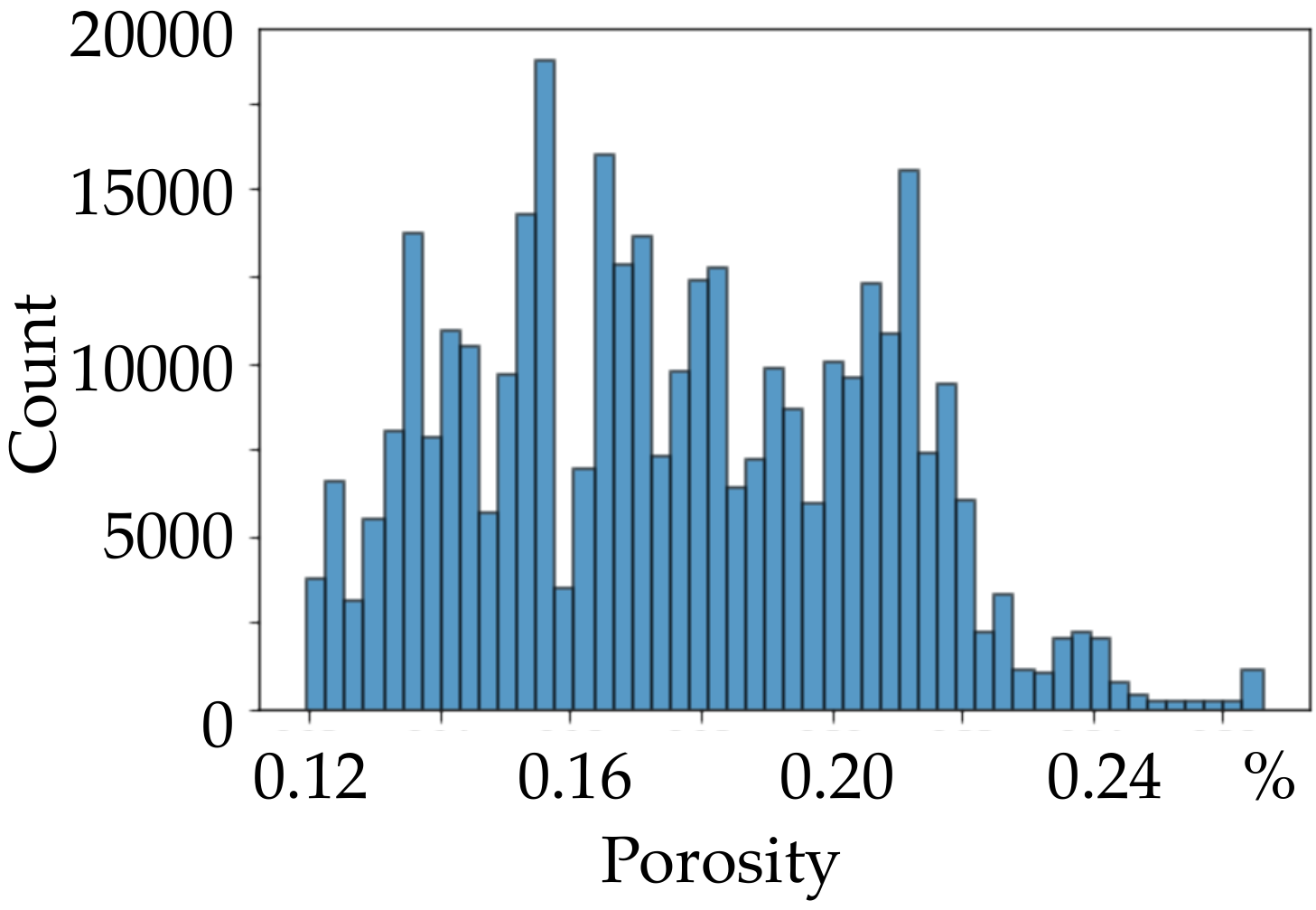}\\a)
\end{subfigure}%
\begin{subfigure}[b]{.55\textwidth}
  \centering
  \includegraphics[width=0.95\linewidth]{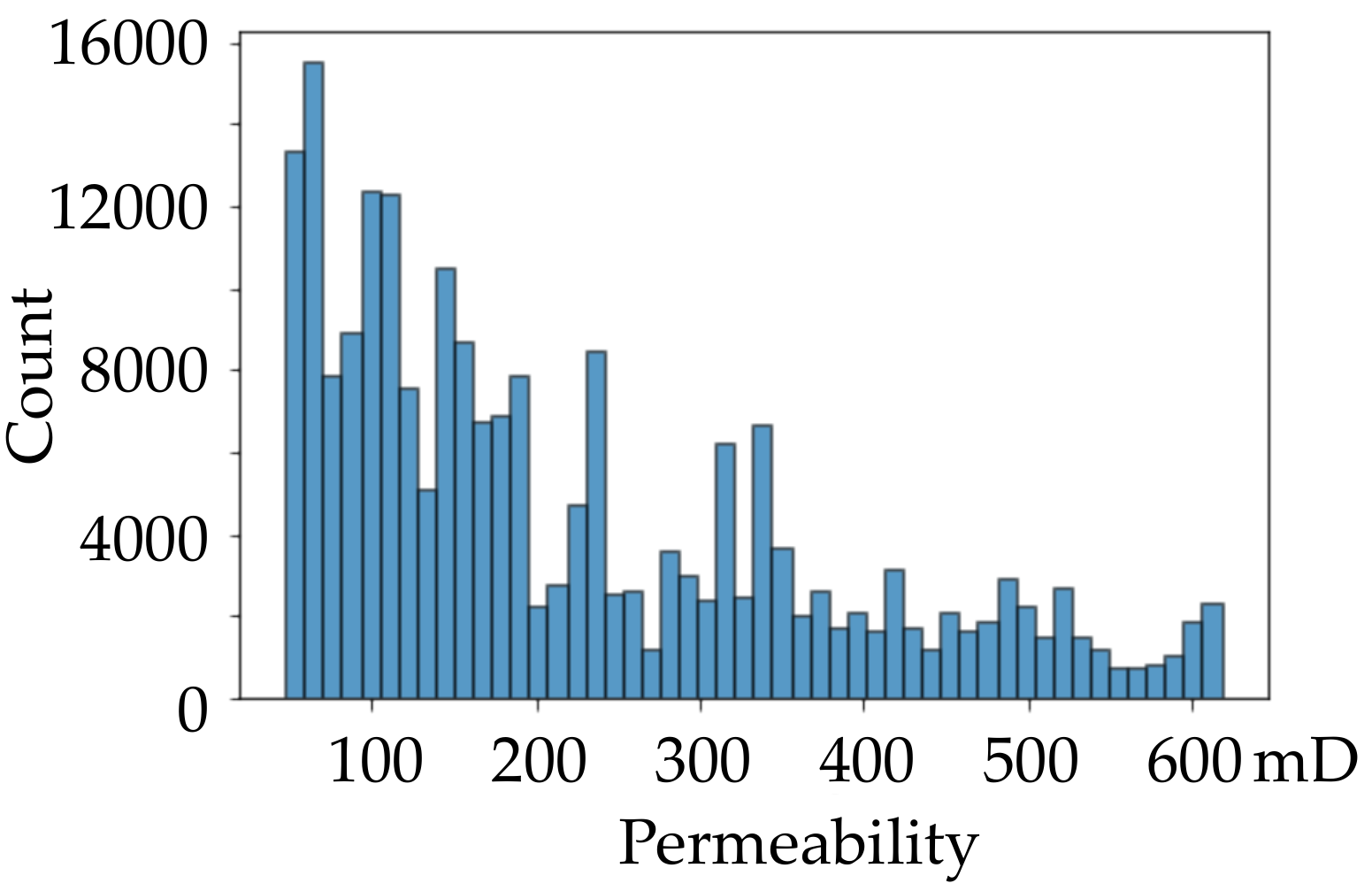}\\b)
\end{subfigure}
\caption{Porosity (a) and permeability (b) values samples from Watt benchmark model.}
\label{fig:fes}
\end{figure}

To the best of our knowledge fluid flow simulation models based on PDE (Partial Differential Equations) remains the industry standard for accurate production forecast. To compare results of RF estimation by the BNs approach, we have chosen a benchmark model provided by Arnold et al. \cite{arnold2013hierarchical}. The Watt field contains a whole range of possible uncertainties from seismic interpretation to fluid parameters, which allow a robust forecast of production levels. Fig. \ref{fig:WOPR} presents production profiles simulated via the ECLIPSE black oil model. Categorical and continuous parameters were sampled from the Watt model in order to compare RF estimation. Fig. \ref{fig:fes} shows distributions for porosity and permeability from well log data in the Watt field. These data were used to build petrophysical models which are discussed in the \ref{RF} section.

\end{document}